\pdfoutput=1

\documentclass[11pt]{article}

\usepackage[final]{ACL2023}
\usepackage{times}
\usepackage{graphicx}
\usepackage{latexsym}
\usepackage{wrapfig}
\usepackage{booktabs}
\usepackage{tabularx}
\usepackage{amsmath}
\usepackage{cleveref}
\usepackage{multirow}

\usepackage[T1]{fontenc}

\usepackage[utf8]{inputenc}

\usepackage{microtype}

\usepackage{inconsolata}

\title{Can Vision-Language Models Evaluate Handwritten Math?}

\newcommand{\bnch}[0]{{\textbf{\textsc{Fermat}}}}

\newcommand{\gptfouro}{{\textsc{GPT-4o}}}
\newcommand{\gptfouromini}{{\textsc{GPT-4o-mini}}}
\newcommand{\geminipro}{{\textsc{Gemini-1.5-Pro}}}
\newcommand{\geminiflash}{{\textsc{Gemini-1.5-FLash}}}
\newcommand{\pixtral}{{\textsc{Pixtral-12B}}}
\newcommand{\llamamini}{{\textsc{Llama-3.2-11B}}}
\newcommand{\llamabig}{{\textsc{Llama-3.2-90B}}}
\newcommand{\phivision}{{\textsc{Phi-3.5-VI}}}
\newcommand{\pixtrallarge}{{\textsc{Pixtral-124B}}}

\definecolor{computational_bg}{HTML}{DAF0F2} 
\definecolor{conceptual_bg}{HTML}{FFEAD9}    
\definecolor{presentation_bg}{HTML}{FBDEDE}    
\definecolor{notational_bg}{HTML}{E0F2E7}   
\definecolor{superficial_bg}{HTML}{EEEAFB}     

\newcommand{\edcot}{{\textbf{\textsc{ED}}}}
\newcommand{\edcotocr}{{\textbf{\textsc{ED+OCR}}}}
\newcommand{\elnoocr}{{\textbf{\textsc{EL}}}}
\newcommand{\elocr}{{\textbf{\textsc{EL+OCR}}}}

\newcommand{\ecnoocr}{{\textbf{\textsc{EC}}}}
\newcommand{\ecocr}{{\textbf{\textsc{EC+OCR}}}}

\author{
 \textbf{Oikantik Nath\textsuperscript{1,2}} \quad
 \textbf{Hanani Bathina\textsuperscript{2,3}} \quad
 \\
 \textbf{Mohammed Safi Ur Rahman Khan\textsuperscript{1,2}} \quad
 \textbf{Mitesh M. Khapra\textsuperscript{1,2}}
 \\
 \textsuperscript{1}Indian Institute of Technology, Madras \quad
 \textsuperscript{2}AI4Bharat \quad
 \textsuperscript{3}Chennai Mathematical Institute
 \\
 \footnotesize{
 \textbf{Correspondence:} \texttt{\{oikantik, miteshk\}@cse.iitm.ac.in, hananib.mds2024@cmi.ac.in}
 }
 \\
\raisebox{-0.3em}{\includegraphics[height=1.1em]{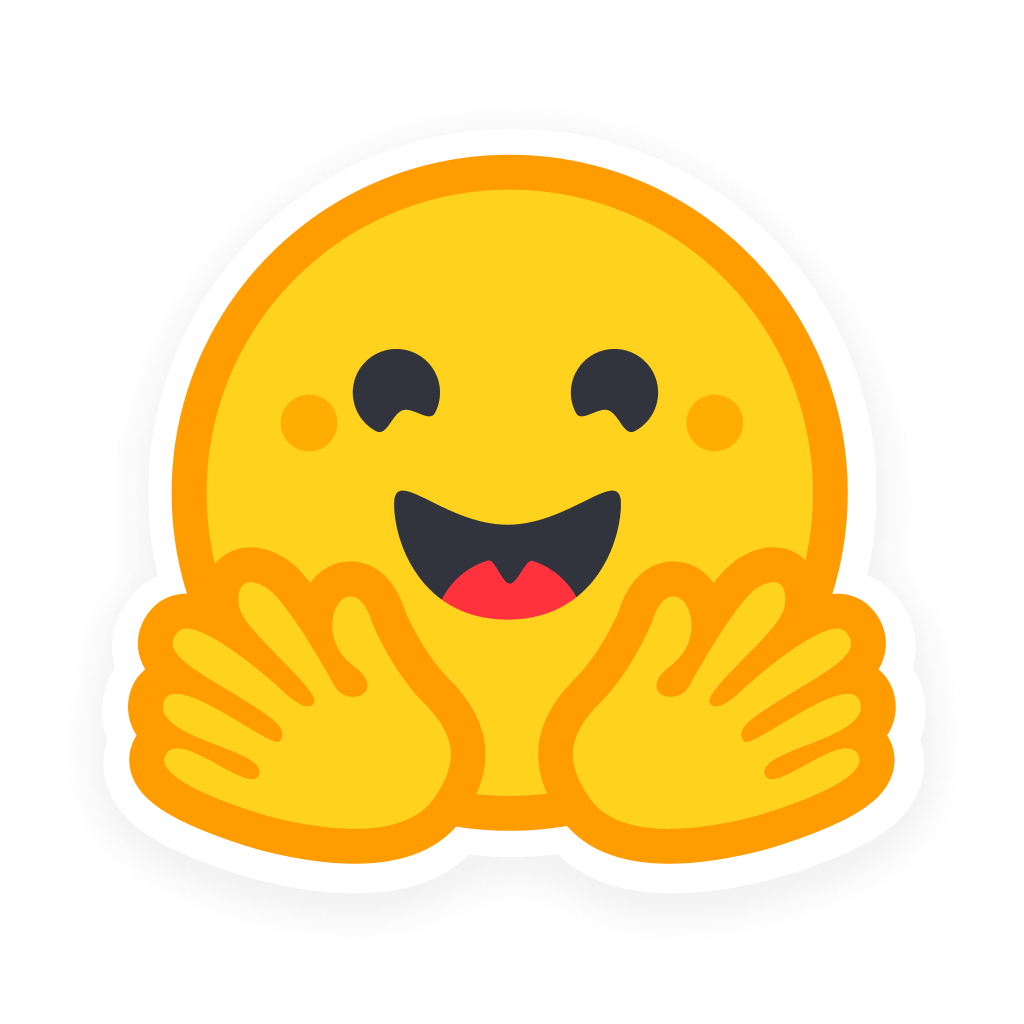}}~\small{\url{https://huggingface.co/datasets/ai4bharat/FERMAT}}
\\
\raisebox{-0.2em}{\includegraphics[height=1em]{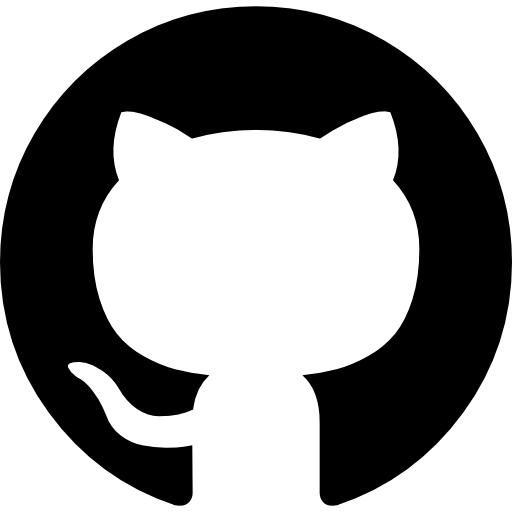}}~\small{\url{https://github.com/AI4Bharat/FERMAT}}
}

\begin{document}
\maketitle

\begin{abstract}
Recent advancements in Vision-Language Models (VLMs) have opened new possibilities in automatic grading of handwritten student responses, particularly in mathematics. However, a comprehensive study to test the ability of VLMs to evaluate and reason over handwritten content remains absent. To address this gap, we introduce {\bnch}, a benchmark designed to assess VLMs' ability to detect, localize and correct errors in handwritten mathematical content. {\bnch} spans four key error dimensions — computational, conceptual, notational, and presentation — and comprises over 2,200 handwritten math solutions derived from 609 manually curated problems from grades 7-12 with intentionally introduced perturbations. Using \bnch~we benchmark nine VLMs across three tasks: error detection, localization, and correction. Our results reveal significant shortcomings in current VLMs in reasoning over handwritten text, with \geminipro~achieving the highest error correction rate (77\%). We also observed that some models struggle with processing handwritten content, as their accuracy improves when handwritten inputs are replaced with printed text or images. \textit{These findings highlight the limitations of current VLMs and reveal new avenues for improvement.} We release \bnch~and all the associated resources in the open-source to drive further research.
\end{abstract}

\section{Introduction}
\label{sections:1_intro}

\begin{figure}[t]
  \centering
  \includegraphics[width=\columnwidth]{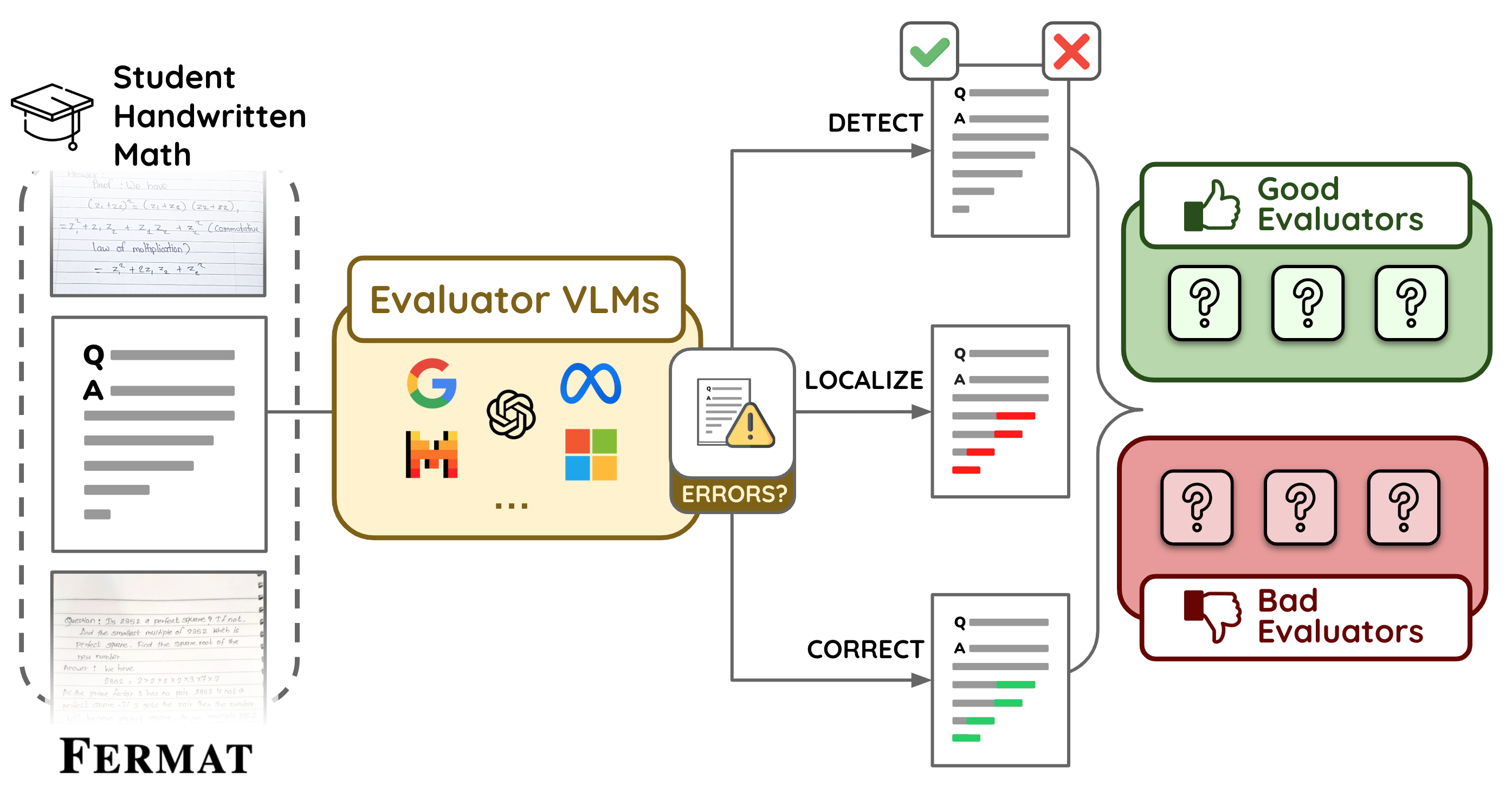}

  \caption{We introduce {\bnch}, a novel multimodal benchmark to evaluate VLMs on their ability to detect, reason about, and assess the correctness of handwritten grade-school level math solutions.}
  \label{fig:cover-img}
\end{figure}

Recent advancements in Large Language Models (LLMs) \citep{ mistral, llama2, qwen2, palm2} and Vision-Language Models (VLMs) \cite{gemini1.5, llama3, gpt4o, qwen2vl, pixtral, llava} have significantly enhanced the ability to interpret both textual and visual data. These developments are driving progress in core language \cite{llmsurvey, llm_se, llm_edu} and vision-language tasks \cite{vlm_survey, multimodal_foundation_models}, with notable advancements in mathematical reasoning and problem-solving \cite{chatgpt_math, math_reasoning_survey, deepseekmath_reasoning, mathcheck, mathprompter, chatgpt_reasoning}. As these models evolve, they are increasingly enabling sophisticated applications in educational tools \cite{llm_survey_edu}, including automated evaluation \cite{autograding, content_creation_evaluation, deduction_under_perturbed_evidence, automated_check_submission}, quiz generation \cite{content_creation_evaluation, auto_q_gen}, and personalized tutoring systems \cite{expert_guided_remediation, intelligent_tutoring, llm_learning_recommendations, llm_knowledge_based_qa}.

One promising application of VLMs is exemplified by OpenAI’s widely referenced demo\footnote{\url{https://www.youtube.com/watch?v=_nSmkyDNulk}}, which demonstrated the potential of such models to evaluate handwritten math content produced by students. This requires a model to accurately understand, identify, and correct potential errors. Although these demonstrations highlight potential, a robust and comprehensive evaluation of VLMs for this task remains lacking. To address this gap, a benchmark analogous to Checklist-based fine-grained assessments for text \citep{checklist, mathcheck, deduction_under_perturbed_evidence} is essential.

To address this need, we introduce {\bnch}, a benchmark to evaluate a VLM's capability in \textbf{F}inding and correcting \textbf{\textsc{er}}rors in handwritten \textbf{\textsc{mat}}hematical content. 
This benchmark enables the evaluation of Vision-Language Models (VLMs) as automatic evaluators for handwritten math responses across four common error axes: (a) computational errors, (b) conceptual misunderstandings, (c) notation errors, and (d) presentation issues. To accomplish this, we first manually curated 609 math problems from grades 7 to 12, along with their correct solutions. We then used a human-in-the-loop approach to introduce targeted perturbations into these correct solutions along the previously defined error axes. Finally, these perturbed solutions were transcribed by more than 40 human annotators to produce handwritten versions. The transcriptions reflect natural variations in handwriting styles, and the captured images reflect differences in lighting, paper types, and overall image quality. The resulting benchmark contains more than 2200 handwritten \textit{erroneous} math solutions and their corresponding correct ``gold'' answers in {\LaTeX} format.

Using \bnch, we evaluate nine VLMs on three core tasks: (a) Error Detection, (b) Error Localization, and (c) Error Correction. Our experiments show that most models struggle with these tasks, with \geminipro~leading with the best performance of 77\% in Error Correction. We also find that providing additional meta-information about the problem type, grade level, error category, etc. improves model performance. Furthermore, our analysis shows that Error Localization accuracy increases when handwritten inputs are replaced with printed images or direct text, highlighting the challenges in processing handwritten content. Overall, these findings highlight key limitations in modern VLMs when processing handwritten mathematical content, emphasizing the need for caution in real-world applications.

\section{Related Work}

\begin{figure*}[!t]
    \centering
    \includegraphics[width=\textwidth]{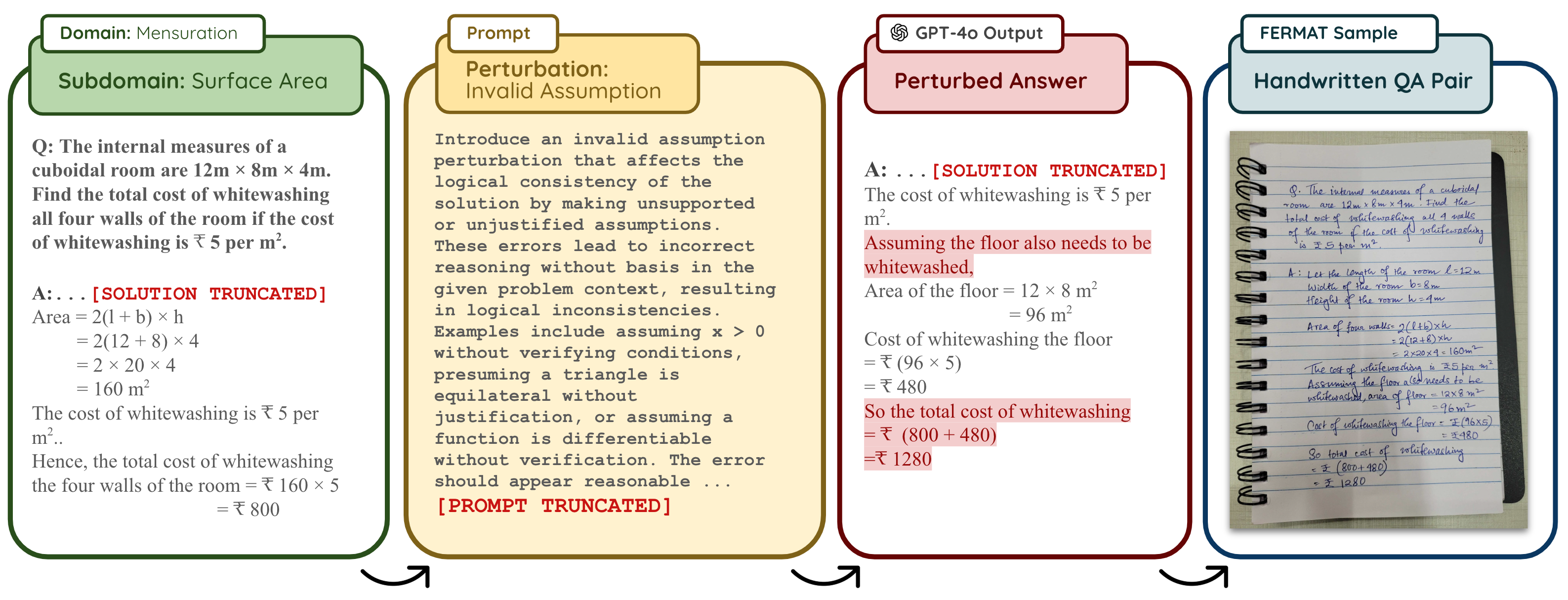}

    \caption{The construction of \bnch~involves four steps: (1) sampling problems with detailed solutions from math domains (\S\ref{problem_set_collection}), (2) defining a perturbation taxonomy (\S\ref{perturbation_taxonomy}), (3) applying perturbations to solutions (\S\ref{perturbed_set_curation}), and (4) transcribing the perturbed QA pairs (\S\ref{hw_transcription}).}
\label{fig:placeholder_with_diversity_axes}
\end{figure*}

\noindent \textbf {Multimodal Evaluations.} The evaluation of VLMs across different multimodal tasks has garnered significant attention in recent works. Prior works~\citep{m3exam, mmmu, examsv, mmmupro, agieval} have introduced multi-disciplinary benchmarks using questions from different competitive exams. Additionally, reasoning benchmarks, including mathematical~\citep{emnlp-2022-lila, mathvista, mathvision} and broader STEM-oriented benchmarks~\citep{olympiadbench}, have been widely explored. While most existing studies evaluate images paired with simple typed text, \citet{ocrbench} and \citet{arxiv_2023_gpt4_expts} investigate OCR capabilities for handwritten text, focusing on single-line mathematical expressions. In contrast, our benchmark includes dense, handwritten, multi-line derivations and complex mathematical notations, hence providing a more rigorous evaluation.

\noindent \textbf{Error Evaluation Abilities of LLMs.} Prior studies~\citep{realmistake, fbi, lema} have explored LLMs' ability to detect textual errors. Some works~\citep{llm_loc_corr, llm_corr_good, deduction_under_perturbed_evidence} highlight that, although LLMs struggle with error detection in mathematical text, they show strong correction abilities. While most research has focused on text-based contexts, a few works~\citep{errorradar, mathcheck} examine multimodal error detection, primarily targeting simple objective errors. In contrast, our benchmark introduces a more realistic evaluation, including multiple variations of a single error type, resulting in a deeper assessment of VLMs’ ability to identify and correct complex multimodal mathematical errors.

\noindent \textbf {\textsc{CheckList}-inspired Work.} The \textsc{CheckList} framework~\citep{checklist} established a systematic approach for evaluating NLP models via behavioral testing. Its principles have been adapted for LLM evaluations, such as FBI~\citep{fbi}, \textsc{MathCheck}~\citep{mathcheck}, and DUPE~\citep{deduction_under_perturbed_evidence}, with a focus on robustness by introducing controlled perturbations in the outputs. Building on this foundation, we introduce a tailored perturbation taxonomy for evaluating handwritten error detection, localization and correction ability of different VLMs.

\section{{\bnch} Benchmark}
\label{sections:3_benchmark}

We present \bnch, a benchmark of 2,244 handwritten solved math problems spanning middle and high school topics, including Arithmetic, Algebra, Mensuration, Geometry, Probability, Statistics, Trigonometry and Calculus. Each solution reflects common mistakes made by students across four different axes: (i) computational errors, (ii) conceptual misunderstandings, (iii) notation errors, and (iv) presentation issues. Additionally, we also include some superficial perturbations that do not render the solution incorrect (e.g., “16 cm” vs. “16.0 cm”). Each instance in {\bnch} comprises a tuple $(Q, I_{hw}, A_{gold})$, where $Q$ represents the question, $I_{hw}$ denotes the image containing the handwritten question and  the \textit{erroneous} solution, and $A_{gold}$ is the original correct solution of $Q$. Both $Q$ and $A_{gold}$ are provided in \LaTeX~to ensure standard uniform representation across different benchmarks. 

The introduced errors are based on well-defined axes of commonly occurring errors designed to rigorously test multimodal reasoning and auto-evaluation capabilities of VLMs. A detailed description of these axes can be found in Table \ref{tab:perturbation_details}. To ensure high standards and sanctity of the benchmark, each instance undergoes multiple stages of manual vetting, from problem-set curation (\S\ref{problem_set_collection}), defining different error categories (\S\ref{perturbation_taxonomy}), creating perturbations (\S\ref{perturbed_set_curation}), to manually transcribing and verifying the perturbed handwritten answers (\S\ref{hw_transcription}). 

\subsection{Problem Set Collection} \label{problem_set_collection}


\noindent \textbf{Initial Data Collection} We first manually collect well-formulated solved problems from widely recognized math textbooks commonly used in grades 7 to 12 curricula. These problems and their solutions are extracted as images from these textbooks, ensuring a diverse representation of core mathematical domains, including Arithmetic, Algebra, Mensuration, Geometry, Probability, Statistics, Trigonometry, and Calculus. This approach ensures comprehensive coverage of foundational concepts across middle and high school levels.

This initial problem set includes only problems with detailed free-form solutions. To enhance the diversity of question formats, we also include multiple-choice questions (MCQs) along with their solutions. These MCQs, sourced from various competitive exams, cover key topics in Quantitative Aptitude, such as profit and loss, time and work, and data interpretation. These topics often involve practical applications of mathematical concepts often underrepresented in standard textbooks.

\noindent \textbf{\LaTeX~Conversion and Verification} After collecting around 850 diverse problem-solution images, we used \gptfouro~to extract the content in {\LaTeX} format. We choose \gptfouro~over standard OCR engines due to its superior capability in handling complex mathematical notations~\citep{Kaltchenko2024AssessingGM} and its ability to give well-formatted outputs. All the extracted \LaTeX~content was then rigorously reviewed by the authors for correctness, resulting in 609 high quality \LaTeX~problem-solution pairs $(Q, A_{gold})$, spanning more than 50 fine-grained topics across the above mentioned 7 domains.




\subsection{Designing the Perturbation Taxonomy} \label{perturbation_taxonomy}

\begin{table}[!t] 
\renewcommand{\arraystretch}{1.1} 
    \footnotesize
    \centering
    \begin{tabular}{lr}

        \toprule
        \textbf{Category} & \textbf{\# Instances} \\
        \midrule
        
        \textbf{\textsc{Total Number of Questions}} & \textbf{\textsc{2,244}} \\
        \quad Free-Form Question-Answer Pairs & 1,814 (82\%) \\
        \quad MCQs with Free-Form Explanations & 430 (18\%) \\
        \midrule
        \textbf{\textsc{Domains (\# Subdomains)}} & \textbf{ } \\
        \quad Algebra (11) & 686 (28.6\%) \\
        \quad Aptitude (1) & 430 (17.9\%) \\
        \quad Arithmetic (13) & 500 (20.9\%) \\
        \quad Calculus (8) & 305 (12.8\%) \\
        \quad Mensuration and Geometry (11) & 260 (10.9\%) \\
        \quad Probability and Statistics (4) & 109 (4.6\%) \\
        \quad Trigonometry (2) & 101 (4.2\%) \\
        \midrule
        \textbf{\textsc{Grade Levels}} & \textbf{\textsc{7 - 12}} \\
        \midrule
        Total Number of Annotators & 43 \\
        Average Annotations per Annotator & 55.6 \\
        \midrule

    \end{tabular}
    \caption{Key statistics of \bnch. Subdomains and perturbation versus grade are detailed in \Cref{appendix:dom_subdom}.}
    \label{tab:dataset_stats}
\end{table}

To reflect common mistakes made by students, we manually designed a comprehensive taxonomy of perturbations specific to our mathematical domains. These perturbations, introduced into correct solutions, are categorized into five broad axes:

\setlength\fboxsep{1.2pt}
\noindent \colorbox{computational_bg}{\textbf{Computational Errors (CO)}}: Errors made in different computations, such as arithmetic mistakes in intermediate or final steps.

\noindent \colorbox{conceptual_bg}{\textbf{Conceptual Errors (CP)}}: Errors made while incorrectly applying concepts, including misinterpretations (e.g., solving for area instead of perimeter) or misuse of identities, like $(a + b)^2 = a^2 + b^2$.

\noindent \colorbox{notational_bg}{\textbf{Notational Errors(NO)}}: Errors made by incorrect usage of symbols, operators, or formulae, such as writing $x^2$ as $x2$ or substituting $+$ for $\times$.

\noindent \colorbox{presentation_bg}{\textbf{Presentation Errors(PR)}}: Clarity or formatting issues, such as providing an answer in fraction form when a decimal is requested, or using inconsistent terminology (e.g., switching between "vector" and "line") that may cause contextual confusion.

\noindent\colorbox{superficial_bg}{\textbf{Superficial Perturbations (SU)}}: Non-impactful errors made by making subtle changes, such as superficially altering variable names ($f(x) = x^2$ to $f(t) = t^2$) or omitting non-essential intermediate steps without affecting solution correctness. These errors evaluate the VLMs' ability to maintain evaluation accuracy despite superficial modifications.

A detailed description of each error axis and perturbations are provided in Table \ref{tab:perturbation_details}. For each of these, the VLM is expected to detect, and correct errors accurately, while ignoring the superficial perturbations that do not affect the solution's validity.

\renewcommand{\arraystretch}{1.1} 
\begin{table*}[t!]
    \centering
    \footnotesize
    \begin{tabular}{lrl}
    
    \toprule
    \textbf{Perturbation Axes} & \textbf{\# Inst} & \textbf{Perturbation Description}  \\ 
    \midrule

    \rowcolor{computational_bg}
    \textbf{\small \textsc{Computational (CO)}} & \textbf{\textsc{$611$}} & \textsc{Calculation \& Propagation Errors}\\
    \textsc{Final Number}  & 156 & Incorrect final answer including digit swaps or misplaced decimals.\\
    \textsc{Intermediate Calculation}  & 100 & Arithmetic calculation errors in intermediate steps.\\
    \textsc{Non-Propagated Step Error} & 80 & Error in intermediate step corrected in subsequent steps. \\ 
    \textsc{Propagated Step Error} & 108 & Error in intermediate step carried forward. \\
    \textsc{Copy Error} & 167 & Copying wrong numbers/expressions from question (e.g., copying \( 45 \) as \( 54 \)). \\ 
    \midrule

    \rowcolor{conceptual_bg}
    \textbf{\small \textsc{Conceptual (CP)}} & \textbf{\textsc{609}} & \textsc{Incorrect Interpretation of Concepts}\\
    \textsc{Theorem Misapplication} & 62 & Applying theorems/identities incorrectly (e.g., using \( \sin^2 \theta + \cos^2 \theta = 0 \)). \\ 
    \textsc{Misinterpret Question} & 145 & Misreading problem requirements such as reporting area instead of volume. \\ 
    \textsc{Invalid Assumption} & 122 & Making assumptions without justification/verification. \\  
    \textsc{Outright Incorrect Fact} & 143 & Stating objectively false information (e.g., a triangle has \textcolor{Maroon}{two} right angles). \\ 
    \textsc{Formula Misuse} & 137 & Incorrectly writing a standard formula (e.g., \( \text{Circle Area: } \textcolor{ForestGreen}{\pi r^2} \rightarrow \textcolor{Maroon}{2 \pi r} \)). \\
    \midrule

    \rowcolor{notational_bg}
    \textbf{\small \textsc{Notational (NO)}} & \textbf{\textsc{255}} & \textsc{Mistakes in Math Symbols \& Operators}\\
    \textsc{Symbol Error} & 81 & Mistakes in symbols/notation (e.g., \( \textcolor{ForestGreen}{x^2} \rightarrow \textcolor{Maroon}{x2} \)). \\ 
    \textsc{Operator Swap} & 115 & Incorrect substitution of operators (e.g., \( \textcolor{ForestGreen}{+} \rightarrow \textcolor{Maroon}{\times} \)). \\ 
    \textsc{Misplaced Parentheses} & 59 & Misplacing parentheses, thus changing the intended order of operations. \\ 
    \midrule

    \rowcolor{presentation_bg}
    \textbf{\small \textsc{Presentation (PR)}} & \textbf{\textsc{429}} & \textsc{Issues in Formatting \& Logical Flow}\\
    \textsc{Format Ignored} & 47 & Ignoring question-specified format (e.g., standard vs scientific notation). \\ 
    \textsc{Terminology Swap} & 25 & Switching inconsistently between terms (e.g., “vector” $\longleftrightarrow$ “line”). \\ 
    \textsc{Logic Disruption} & 101 & Presenting steps out of logical order (e.g., final answer used in earlier steps). \\ 
     \textsc{Contextual Swap} & 43 & Contextually similar but incorrect term substitution (e.g.,  \textcolor{ForestGreen}{circle} $\rightarrow$ \textcolor{Maroon}{ellipse}). \\ 
    \textsc{Variable Misnaming} & 67 & Swapped variables (e.g., swapping \( a \) and \( b \) in a quadratic formula). \\ 
    \textsc{Incorrect Units} & 146 & Reporting with wrong units (e.g., length in kg instead of m). \\ 
    \midrule

    \rowcolor{superficial_bg}
    \textbf{\small \textsc{Superficial (SU)}} & \textbf{\textsc{340}} & \textsc{Modifications Without Impacting Correctness}\\
    \textsc{Superficial Var Change} & 100 & Superficially changing variable names (e.g., \( f(x)=x^2 \rightarrow f(t)=t^2 \)). \\
    \textsc{Step Omission} & 81 & Skipping non-essential intermediate steps. \\
    \textsc{Irrelevant Info} & 159 & Including unnecessary information (e.g., adding unrelated discussions). \\
    \midrule
    \textbf{\normalsize \textsc{Total Instances}} & \textbf{\textsc{2244}} &\\
    \midrule
    \end{tabular}
    \caption{Overview of perturbation categories with descriptions for perturbation. Correct original text is highlighted in \textcolor{ForestGreen}{green}, while perturbed text is highlighted in \textcolor{Maroon}{red}.}
    \label{tab:perturbation_details}
\end{table*}

\subsection{Human-In-The-Loop Perturbation Generation}

\label{perturbed_set_curation}

Based on the perturbation taxonomy (\S\ref{perturbation_taxonomy}), a subset of relevant perturbations is manually selected for each math domain. For each problem curated in (\S\ref{problem_set_collection}), a domain-specific perturbation is applied using \textsc{GPT-4o}, denoted as $f(\cdot)$, by prompting it with the \LaTeX~question $Q$ and correct solution $A_{gold}$. This process is represented as $f(P, X_{P}, Q, A_{gold}) \rightarrow (exp, A_{pert})$, where $P$ is the chosen perturbation, $X_{P}$ represents instructions for inducing the perturbation along with three in-context examples, $A_{pert}$ is the perturbed solution, and $exp$ explains the introduced perturbation. This process is repeated until all problems undergo the relevant perturbations within its domain's subset, ensuring comprehensive coverage.

While \gptfouro~generally produces the intended perturbations, occasional inconsistencies are observed, such as deviations from the specified perturbation, irrelevant modifications, misaligned reasoning, or unchanged answers despite correct reasoning. To address these issues, all perturbed answers $(A_{pert})$ are manually verified by the authors to ensure that intended perturbation is correctly applied and that the reasoning aligns with it. During this review process, the induced perturbations are further classified as true errors or superficial changes. Further details of this are provided in \Cref{appendix:PQCS}.


\subsection{Handwritten Transcription with Manual Verification} \label{hw_transcription} 

We engaged a team of 43 annotators from diverse demographic backgrounds to manually transcribe each perturbed answer $A_{pert}$.  Annotators were instructed to use various paper types and colored pens or inks. The handwritten questions and solutions were captured using mobile phone cameras by the annotators and subsequently uploaded to a centralized portal. This process ensured a diverse benchmark, reflecting a wide range of handwriting styles, paper types, and lighting conditions. As each problem underwent multiple perturbations, the dataset effectively simulates exam-like scenarios where students encounter similar questions but make distinct mistakes in their responses.



Each image $I_{hw}$ was then manually verified by the authors to ensure correct replication of the intended perturbation. During this verification, we recorded additional metadata such as handwriting legibility, image orientation, and overall image quality for each $I_{hw}$. A custom validation tool was developed to streamline this review and annotation process. Detailed statistics on \bnch~ are provided in \Cref{tab:dataset_stats}, and further details on the verification tool in \Cref{appendix:AQCS}.

\section{Evaluation Setup}
\label{sections:5_eval_setup}
In this section, we outline the different tasks on which we evaluate different VLMs on \bnch. Each VLM, denoted by $f(\cdot)$, takes as input a handwritten answer $I_{hw}$ (\S\ref{hw_transcription}) and a prompt $P_{x}$ specific to a task $x$. Detailed prompts for all tasks are provided in \Cref{appendix:specific_prompts}. We propose three tasks of increasing difficulty: (i) Error Detection (\S\ref{err_det}), (ii) Error Localization (\S\ref{err_loc}), and (iii) Error Correction  (\S\ref{err_corr}). For each task, we evaluate multiple strategies, all using a Chain-of-Thought (COT)~\citep{wei2022chain} method, by asking the VLM to provide a step-by-step reasoning before giving its answer. 



\subsection{Error Detection} 
\label{err_det}

In this task, the VLM $f(\cdot)$ is prompted to detect the error in the given handwritten image $I_{hw}$ and give a binary output indicating the presence of an error along with its reasoning.


\paragraph{\edcot:} In this strategy, the VLM, $f (\cdot)$ is directly provided with a handwritten image $I_{hw}$ and a prompt ($P_{ED}$) to detect the error and output a binary value ($True/False$), indicating the presence of an error in the solution and a reasoning ($exp$) for the same. We denote this formally  as $f (P_{ED}, I_{hw}) \rightarrow (exp, True/False)$.
  
\paragraph{\edcotocr:} In this strategy, we decompose the task into two steps, where first the VLM, $f(\cdot)$, is provided  with the handwritten image $I_{hw}$ and  prompt ($P_{OCR}$) to perform OCR and convert the handwritten content into \LaTeX~format. Next, the same VLM, $f(\cdot)$, is prompted with the resulting \LaTeX~text, to detect the error and output a binary value ($True/False$) along with the reason. This is formally denoted as $f(P_{ED}, f(P_{OCR}, I_{hw})) \rightarrow (exp, True/False)$.

\subsection{Error Localization}
\label{err_loc}

In this task, the VLM, $f(\cdot)$, is prompted to accurately localize the error in the given handwritten image $I_{hw}$, by identifying the specific line where the error occurs and providing reasoning for its decision. If no error is present, then the model is asked to output ``NA'' (Not Applicable). This task is more challenging than error detection (ED) (\S\ref{err_det}) since the VLM must perform both error detection and localization simultaneously. 

\paragraph{\elnoocr:} In this strategy, the VLM, $f(\cdot)$ is directly given a handwritten image $I_{hw}$ along with a prompt ($P_{EL}$) to localize the error, if present, in the image. The VLM describes the specific line(s) containing the error(s) and provides an explanation. Formally, this is represented as  $f(P_{EL}, I_{hw}) \rightarrow (exp, text_{loc}/NA)$.



\paragraph{\elocr:} Similar to the \edcotocr~strategy discussed in Sec \S\ref{err_det}, the VLM $f(\cdot)$ is first prompted to perform OCR on the given handwritten image $I_{hw}$ and then asked to localize the error in the output \LaTeX~text by describing the specific line(s) containing the error(s). This is formally denoted as  $f(P_{EL}, f(P_{OCR}, I_{hw})) \rightarrow (exp, text_{loc}/NA)$.


\subsection{Error Correction }
\label{err_corr}

In this task, the VLM $f(\cdot)$ is prompted to correct any errors found in a given handwritten image $I_{hw}$ and output the entire corrected solution in \LaTeX~format. If no error is present, the VLM is asked to output ``NA''. This is the most challenging of the three tasks, as the VLM must perform error detection, localization, and correction in a single step.


\paragraph{\ecnoocr:} In this strategy, the VLM, $f(\cdot)$, is directly given the handwritten image $I_{hw}$ along with the prompt ($P_{EC}$) to correct any errors. If errors are detected, the model outputs the entire corrected solution $A_{corr}$, otherwise, it returns ``NA'' to indicate the solution is already correct. Since a problem can often be solved in multiple different ways to reach the final answer, the model is allowed to explore all possible ways to generate the correct answer to the problem. The error correction strategy is formally denoted as  $f(P_{EC}, I_{hw}) \rightarrow (exp, A_{corr} / NA)$.

\paragraph{\ecocr:} Similar to the strategies discussed in Sec \S\ref{err_det} and \S\ref{err_loc}, the VLM $f(\cdot)$ is first prompted to perform OCR on the given image and then prompted to give the entire corrected answer or ``NA'' if no error is found. Formally, we represent this process as $f(P_{EC}, f(P_{OCR}, I_{hw})) \rightarrow (exp, A_{corr} / NA)$.



\subsection{Cascaded Setup}
\label{cascaded}

\begin{figure}[!t]
    \centering
    \includegraphics[width=0.6\linewidth]{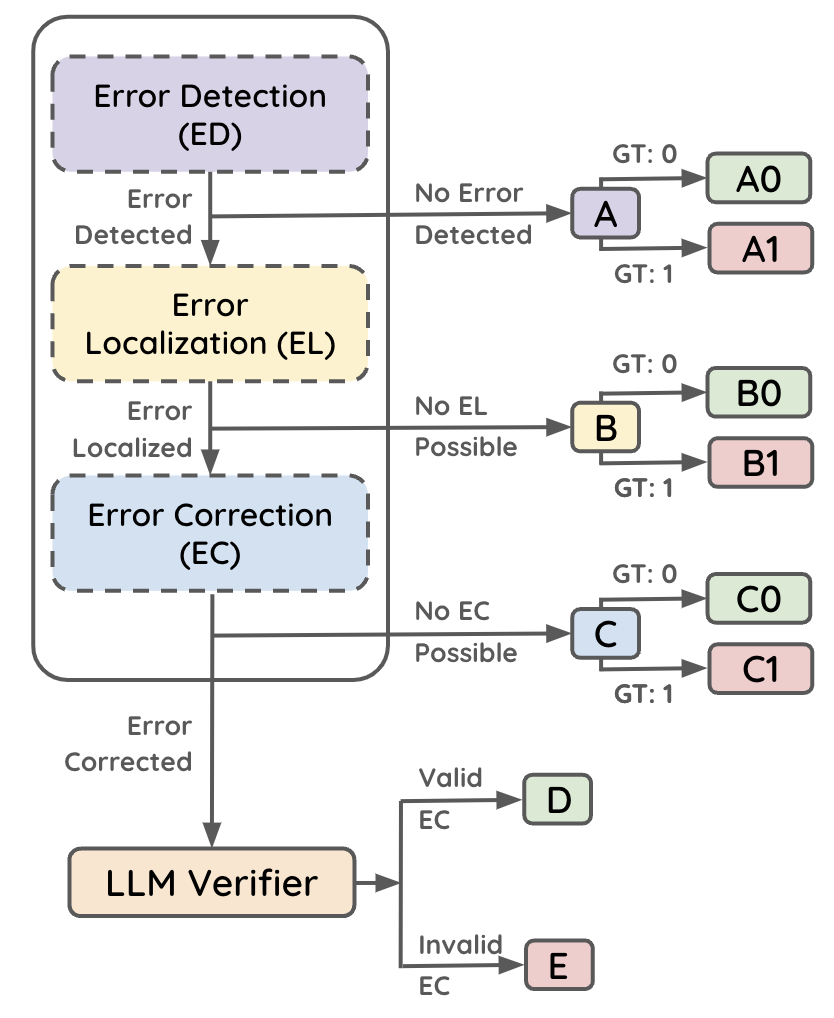}
    \caption{Cascaded black-box evaluation setup, as described in \S\ref{cascaded}.$GT$ denotes Ground Truth. The total number of correctly evaluated \bnch~samples in this setup is represented by the summation of $A0$, $B0$, $C0$, and $D$.}
    \label{fig:cascaded}
\end{figure}

In the above setups, each of the three tasks was performed independently. Here, we evaluate a cascaded setup where these tasks are executed sequentially, as shown in \Cref{fig:cascaded}. In this approach, the VLM $f(\cdot)$ first performs error detection as outlined in \edcot~(\S\ref{err_det}). For images identified as containing errors, error localization is then performed using \elnoocr~(\S\ref{err_loc}). Finally, for images with localized errors, the error correction step is executed based on the method described in \ecnoocr~(\S\ref{err_corr}). Unlike previous setups, the output of each stage is passed as input to the next. For example, during error correction, the VLM is provided with both the original image and the localized error line(s) from the previous step to improve accuracy. The cascaded setup aims to achieve precise error correction by leveraging the context generated at each stage. Formally, this process can be represented as $f(P_{EC}, I_{hw}, f(P_{EL}, I_{hw}, f(P_{ED}, I_{hw}))) \rightarrow (exp, A_{corr}/NA)$.

\begin{table*}[!t]
\footnotesize
\centering
\begin{tabular}{l|@{}|c|c|c|c|c|c|@{}|c}
\toprule
 & \multicolumn{6}{c|@{}|}{\textbf{Non-cascaded}} & \textbf{Cascaded} \\ \cmidrule{2-8} 
\textbf{Models}                                 & \textbf{ED} & \textbf{ED+OCR} & \textbf{EL}   & \textbf{EL+OCR} & \textbf{EC}   & \textbf{EC+OCR} & \textbf{\edcot}~{\raisebox{0.1em}{\includegraphics[height=0.5em]{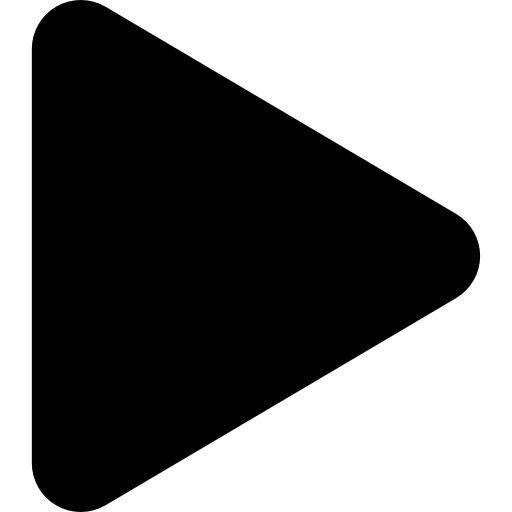}}}~\elnoocr~{\raisebox{0.1em}{\includegraphics[height=0.5em]{figures/right_arrow.png}}}~\textbf{\ecnoocr} \\ \cmidrule{2-8} 
& \textbf{BACC}    & \textbf{BACC}      & \textbf{ACC}  & \textbf{ACC}    & \textbf{ACC}  & \textbf{ACC}    & \textbf{ACC}                                 \\
\midrule
{\raisebox{-0em}{\includegraphics[height=0.9em]{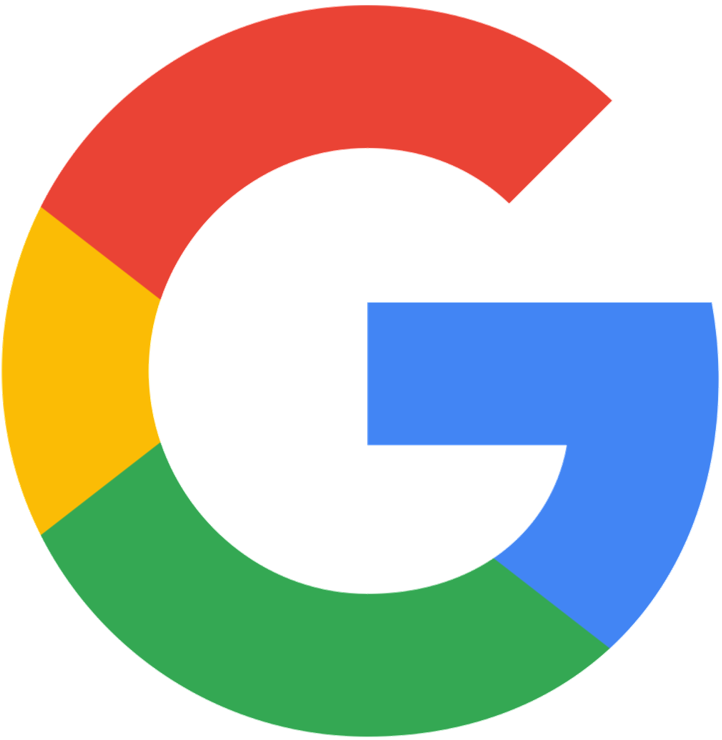}}}~\geminipro          & 0.63                   & \textbf{0.67}             & 0.43          & \textbf{0.56}   & \textbf{0.76} & \textbf{0.77}   & 0.50                                         \\
{\raisebox{-0em}{\includegraphics[height=0.9em]{figures/google.png}}}~\geminiflash        & 0.60                     & 0.62             & 0.39          & 0.51            & 0.70          & 0.72            & 0.46                                         \\

{\raisebox{-0em}{\includegraphics[height=0.9em]{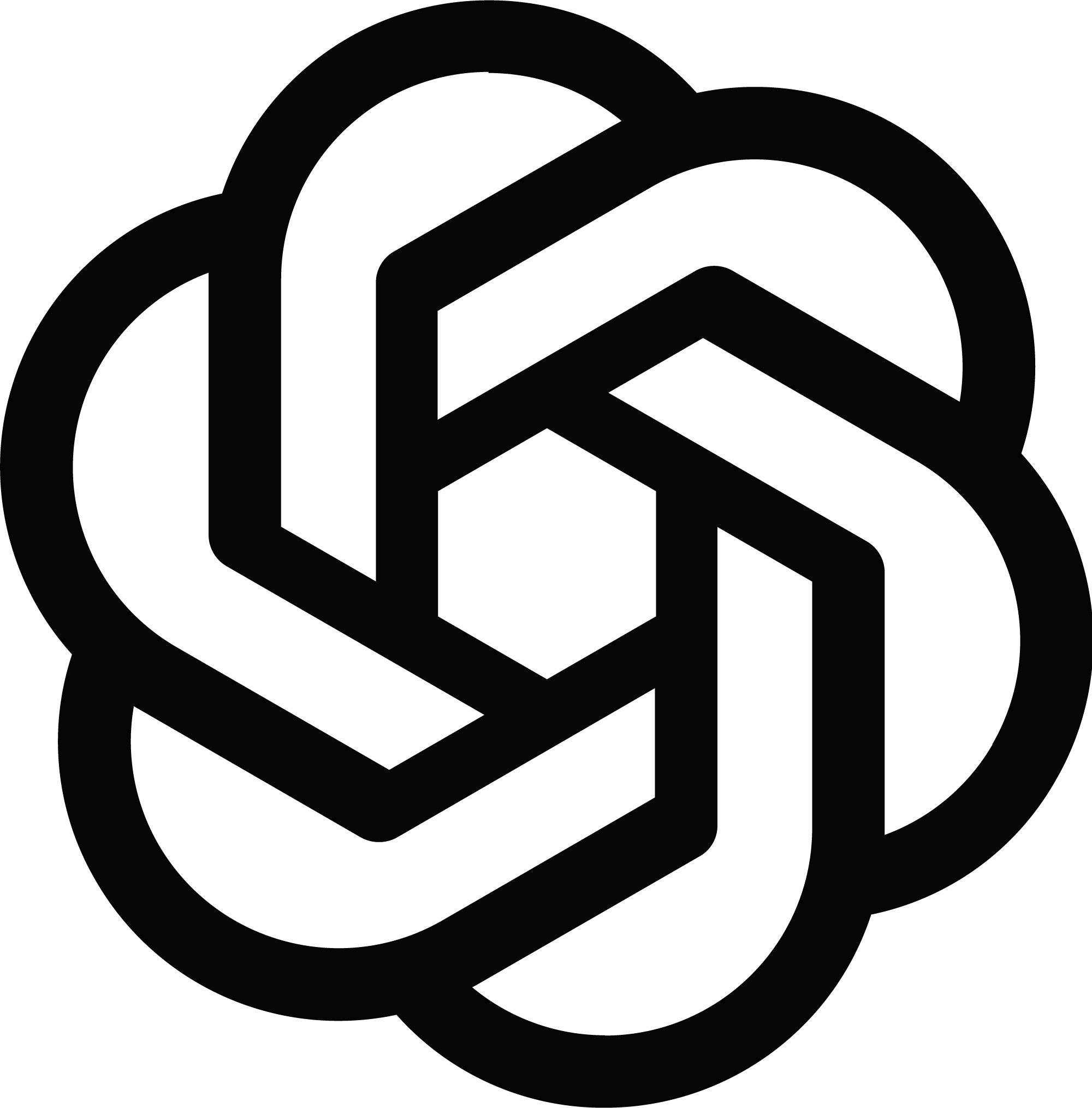}}}~\gptfouro                   & \textbf{0.65}                     & 0.64             & \textbf{0.45}          & 0.50            & 0.66          & 0.71            & 0.45                                         \\
{\raisebox{-0em}{\includegraphics[height=0.9em]{figures/openai.png}}}~\gptfouromini              & 0.55  & 0.57             & 0.44 & 0.45            & 0.56          & 0.58            & \textbf{0.51}                                \\

{\raisebox{-0em}{\includegraphics[height=0.5em]{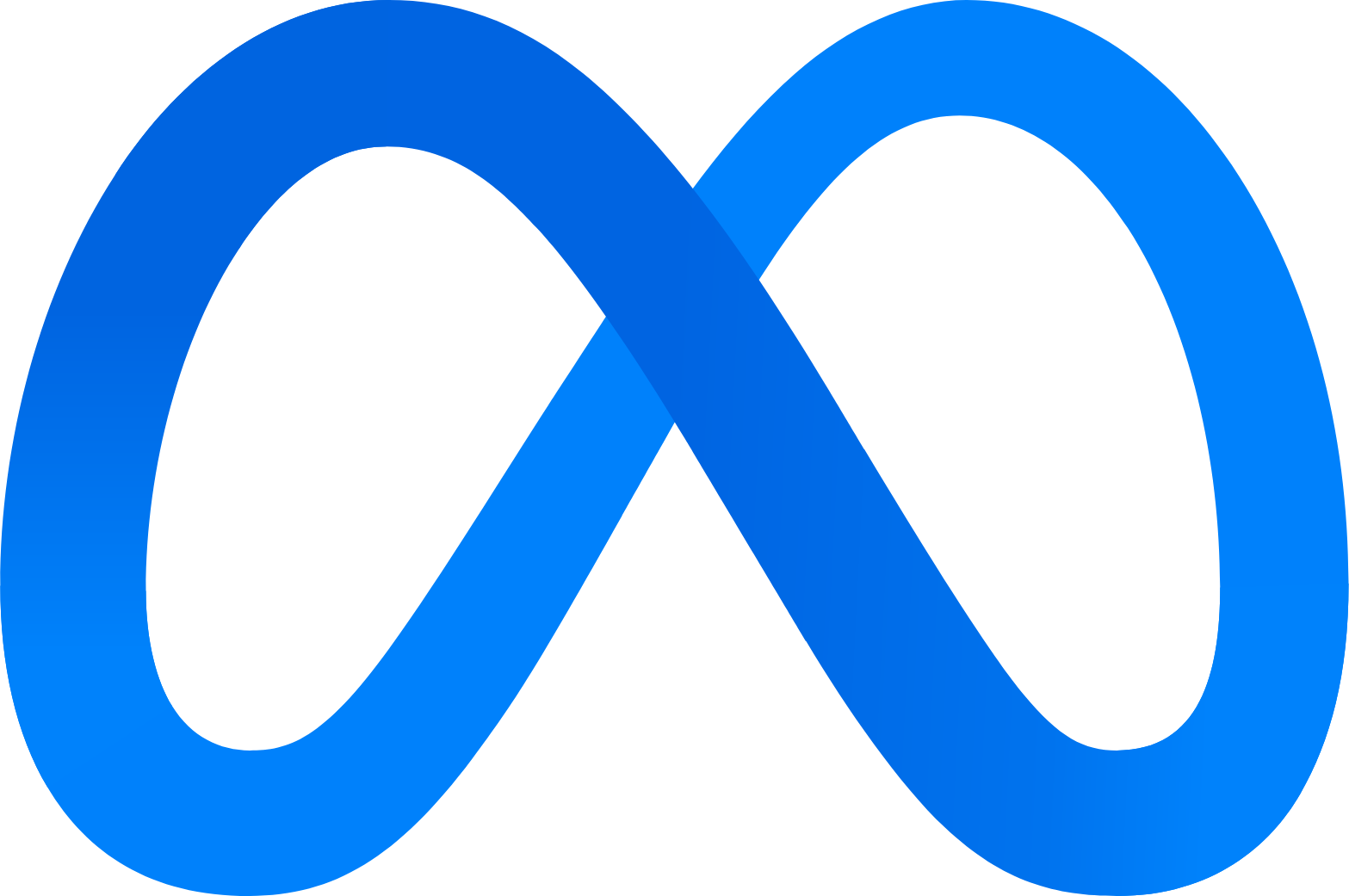}}}~\llamabig               & 0.52                  & 0.62             & 0.18          & 0.41            & 0.25          & 0.57            & 0.31                                         \\
{\raisebox{-0em}{\includegraphics[height=0.5em]{figures/meta.png}}}~\llamamini               & 0.50            & 0.52             & 0.14          & 0.27            & 0.21          & 0.38            & 0.20                                         \\

{\raisebox{-0em}{\includegraphics[height=0.8em]{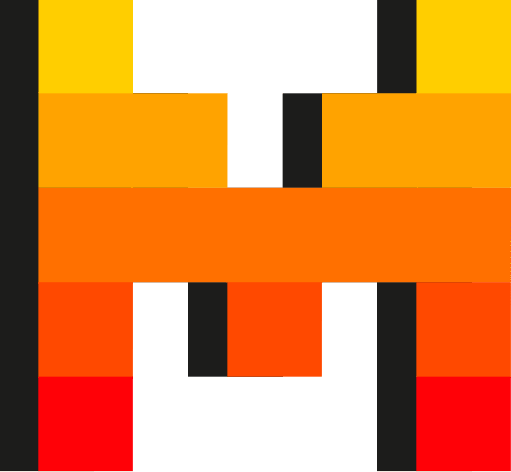}}}~\pixtrallarge            & 0.52                   & 0.59             & 0.24          & 0.40            & 0.46          & 0.56            & 0.26                                         \\
{\raisebox{-0em}{\includegraphics[height=0.8em]{figures/mistral.png}}}~\pixtral             & 0.51        & 0.55    & 0.24          & 0.27            & 0.30          & 0.34            & 0.32                                         \\
{\raisebox{-0em}{\includegraphics[height=0.8em]{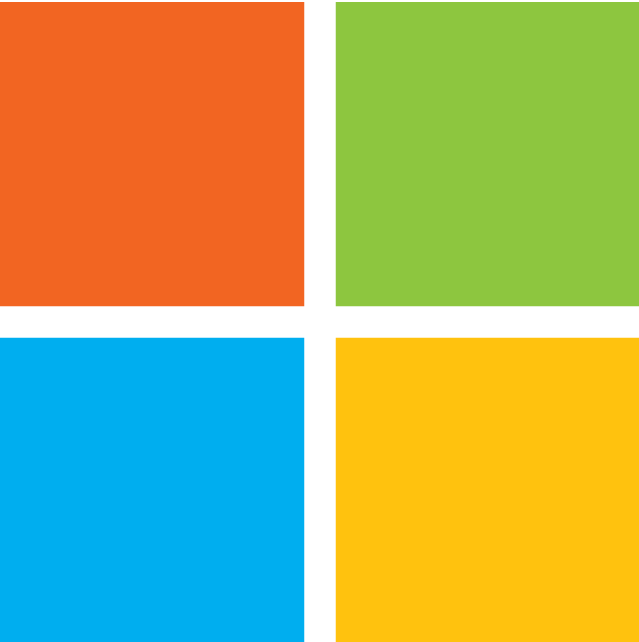}}}~\phivision                & 0.52               & 0.51             & 0.06          & 0.09            & 0.15          & 0.12            & 0.11    \\
\bottomrule
\end{tabular}
\caption{Performance comparison of VLMs in cascaded and non-cascaded settings on {\bnch} across different evaluation strategies. Metrics include Balanced Accuracy (\textbf{BACC}) for error detection, and Accuracy (\textbf{ACC}) for error localization and correction. Higher values ($\uparrow$) indicate better performance.}
\label{tab:eval_non_cascaded}
\end{table*}

\subsection{LLM as an Evaluator}
\label{llm_evaluator}

Error localization (\S\ref{err_loc}) and correction (\S\ref{err_corr}) are inherently subjective tasks, as multiple valid solutions can exist. While human evaluation remains the gold standard for VLM assessment, it is costly and time-intensive. To address this, we use Large Language Models (LLMs) as automated evaluators, following recent advancements~\citep{mtbench, chiang-lee-2023-large}. For localization, the LLM checks if errors are correctly identified, and for correction, it verifies the accuracy of the corrected solution. We use \gptfouro~as our Evaluator LLM due to its widespread use as an evaluator.

To validate the reliability of our \gptfouro-based Evaluator LLM, we conducted a study on 464 randomly sampled task outputs from four VLMs: {\gptfouro}, {\llamamini}, {\pixtral}, and {\phivision}. Graduate students were independently tasked with assessing the VLM outputs to determine their accuracy. We then compared these human judgments with the evaluations produced by our Evaluator LLM and found a \textit{94\%} average agreement between the two. Given this strong alignment with human evaluations, we opted to use our \gptfouro~based Evaluator LLM as a faster but equally reliable alternative to the expensive and time-consuming human evaluations for all subsequent experiments. The prompts used for our Evaluator LLM as well as details about the human verification are provided in \Cref{appendix:human_vs_llm} and \ref{appendix:specific_prompts}.


\section{Experiments}
\label{sections:4_expts}


We evaluate nine popular VLMs, including both closed-proprietary and open-sourced models as listed in Table \ref{tab:eval_non_cascaded} on \bnch. For each task, we ensure consistent evaluation by using identical prompts across all models and setting the sampling temperature to zero to maintain reproducibility. Similarly, for the Evaluator LLM, we use \gptfouro~with a temperature of zero. Detailed prompts for all the experiments are provided in \Cref{appendix:specific_prompts}.


For the Error Detection task (\S\ref{err_det}), we report the model performance using Balanced Accuracy, which accounts for the class imbalance by averaging the \textit{sensitivity} (true positive rate) and \textit{specificity} (true negative rate). Ground truth labels are defined as 0 for Superficial Perturbations (SU) (\S\ref{perturbation_taxonomy}) and 1 for all other error types. We report Balanced Accuracy instead of the standard F1 score since it gives equal importance to both positive and negative labels, whereas the F1 score ignores the true negatives altogether. We provide additional information regarding F1 and Accuracy scores in \Cref{appendix:acc_fp}. For the Error Localization (\S\ref{err_loc}) and Error Correction (\S\ref{err_corr}) tasks, we report Accuracy, which we define as the proportion of times the Evaluator LLM (\S\ref{llm_evaluator}) determines that the VLM has done an accurate job. 

\subsection{How do different VLMs perform?} \label{eval_noncascaded}
We present the main results of our tasks in Table \ref{tab:eval_non_cascaded}. Overall, all models face challenges in the core tasks of \bnch, with \gptfouro~and \geminipro~consistently leading across all tasks. \gptfouro~demonstrates superior performance in the \edcot~and \elnoocr~tasks, while \geminipro~achieves the best results in the remaining tasks. Most models perform well on the Error Detection task, but performance declines significantly as task complexity increases for Localization and Correction. A detailed analysis of this trend is provided in \Cref{tab:acc_fp}. We also observe that introducing an explicit OCR step, improves performance for certain models. Notably, \pixtrallarge ~and \llamabig ~show large gains, which can be attributed to stronger handwriting OCR capabilities compensating for weaker multimodal reasoning. By contrast, models with strong multimodal understanding, such as \gptfouro~and \geminipro, gain marginal benefits from the OCR step, suggesting they rely less on textual signals and are better at jointly interpreting visual and textual content.

\subsection{How do VLMs perform in the Cascaded Approach?} \label{eval_cascaded}


We evaluate all models in the cascaded setup described in \S\ref{cascaded}. As shown in the last column of Table \ref{tab:eval_non_cascaded}, decomposing the Error Correction task into sequential steps leads to a significant performance drop across models, including \gptfouro, \geminiflash, \geminipro, and \pixtrallarge. This decline is primarily attributed to the cautious error detection behavior of these models (discussed in \Cref{tab:acc_fp}), which results in a large proportion of images being filtered out during the initial stage (\edcot). A comprehensive breakdown of intermediate and final outputs for each VLM in the cascaded setup is provided in \Cref{appendix:vlm_cascaded_setup}.

\subsection{Does more information help VLMs?} 
\label{levels_expt}

We conducted an ablation study to evaluate whether providing additional information about the error type improves model performance. Four settings with increasing levels of information were designed: \textbf{L1} (basic context, including grade, math domain, and subdomain), \textbf{L2} (L1 + descriptions of \textit{all} perturbations specific to that domain + some examples of perturbations), \textbf{L3} (L1 + specifying the \textit{exact} perturbation category that was applied), and \textbf{L4} (L3 + a sample \textit{erroneous} solution accompanied by an explanation of the mistake). As shown in \Cref{tab:levels}, performance consistently improves with the addition of more detailed information, indicating that increasing error context facilitates better Error Detection. Prompts designed for this study are provided in \Cref{appendix:specific_prompts}.

We note that while this experiment provides valuable insights from an ablation perspective, incorporating such detailed information may be challenging in practical scenarios. For example, if a teacher is required to specify the exact error type in a solution, they might find it more practical to evaluate the solution directly without relying on a VLM.


\setlength{\tabcolsep}{5pt} 
\renewcommand{\arraystretch}{1.1} 
\begin{table}[t]
\centering
\small
\begin{tabular}{lccccc}
\toprule
\textbf{Model} & \textbf{Base} & \textbf{L1} & \textbf{L2} & \textbf{L3} & \textbf{L4} \\
\midrule
\gptfouro & 0.658 & 0.670 & 0.676 & 0.691 & 0.702 \\
\bottomrule
\end{tabular}
\caption{\textbf{BACC} (Balanced Accuracy) scores of {\gptfouro} on the error detection task under increasing levels of helpful contextual information included in the prompt. Higher scores indicate better performance.}
\label{tab:levels}
\end{table}




\subsection{How much does handwriting affect model performance on \bnch~?}
\begin{figure}[!t]
    \centering
    \includegraphics[width=1.0\linewidth]{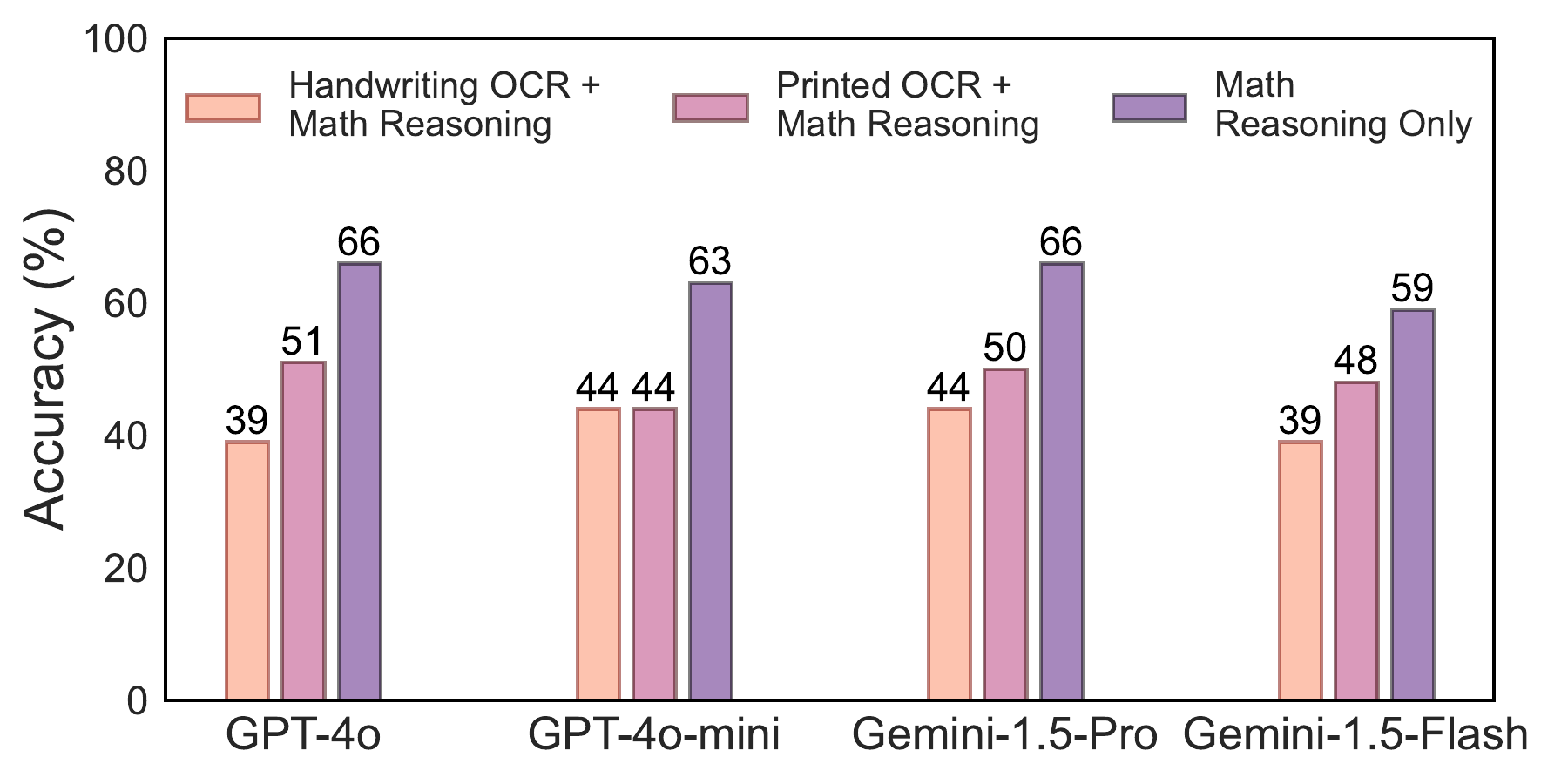}
    \caption{Performance of VLMs on the error localization task across various benchmark settings. Higher scores ($\uparrow$) indicate better performance.}
    \label{fig:vlm_ablation_performance}
\end{figure}


We hypothesize that weaker handwriting recognition capabilities in some models (\Cref{tab:eval_non_cascaded}) may impair their ability to identify and correct mistakes. To test this, we conduct two studies to isolate reasoning abilities from visual processing. First, we replace handwritten images with printed \LaTeX~rendered images from the ($Q$, $A_{perturb}$) pairs (\S\ref{perturbed_set_curation}). Second, we eliminate images entirely, providing direct \LaTeX~text inputs for $Q$ and $A_{perturb}$. As shown in \Cref{fig:vlm_ablation_performance}, performance improves consistently as visual complexity is reduced. The largest gains occur when switching to text input while replacing handwritten images with printed \LaTeX~still offers small benefits on an average. These results highlight the challenges of processing handwritten content and reinforce \bnch's rigor as a benchmark for evaluating both reasoning and visual understanding in VLMs.








\section{Conclusion}
\label{sections:conclusion}


We introduce {\bnch}, a comprehensive benchmark to assess Vision-Language Models (VLMs) on their ability to detect, localize, and correct errors in handwritten mathematical content. By spanning four critical error dimensions — computational, conceptual, notational, and presentation — and curating over 2,200 perturbed handwritten solutions from 609 math problems (grades 7-12), {\bnch} provides a robust evaluation framework. Our analysis of nine prominent VLMs reveals key limitations in their reasoning over handwritten content. While \geminipro~achieves the highest error correction rate (77\%), we find that smaller models often struggle. Our findings also highlight the challenges posed by handwritten content, as models perform better with printed images or text inputs. By releasing {\bnch} and all associated resources as open-source, we hope that this fosters further research on evaluating and enhancing the capabilities of VLMs for real-world applications.

\section*{Limitations}
While we have compiled a comprehensive list of perturbation categories, we acknowledge that it may not be exhaustive, leaving room for further expansion. Our benchmark primarily focuses on school-level mathematics questions, with more advanced topics and question types left for future work. Additionally, we do not explore complex multi-agent approaches for error detection, instead limiting our study to single or dual LLM calls.

\section*{Ethics Statement}
Annotators who participated in the annotation and/or verification task are paid a competitive monthly salary to help with the tasks. The salaries were determined based on the qualification and the prior experience working on similar tasks and adhering to the norms of the government of our country. The annotators were made aware that the datasets will be publicly released. The annotated datasets have no personally identifying information. The datasets used in this paper will be made available under permissible licenses, and we adhere strictly to their intended usage, maintaining compliance with licensing requirements. Additionally, all the code used for our evaluations and perturbation generation will be made publicly available under the MIT License. We only used AI Assistants for assistance purely with the language of the paper, e.g., paraphrasing, spell-checking, or polishing the author’s original content, without suggesting new content

\bibliography{main}
\bibliographystyle{acl_natbib}
\newpage
\section*{Appendix}

\appendix

\section{Additional details of \bnch}
\label{appendix:dom_subdom}

\subsection{Distribution of Math Domains and Perturbation Domains in \bnch}
\begin{figure}[h]
    \centering
    \includegraphics[width=0.9\linewidth]{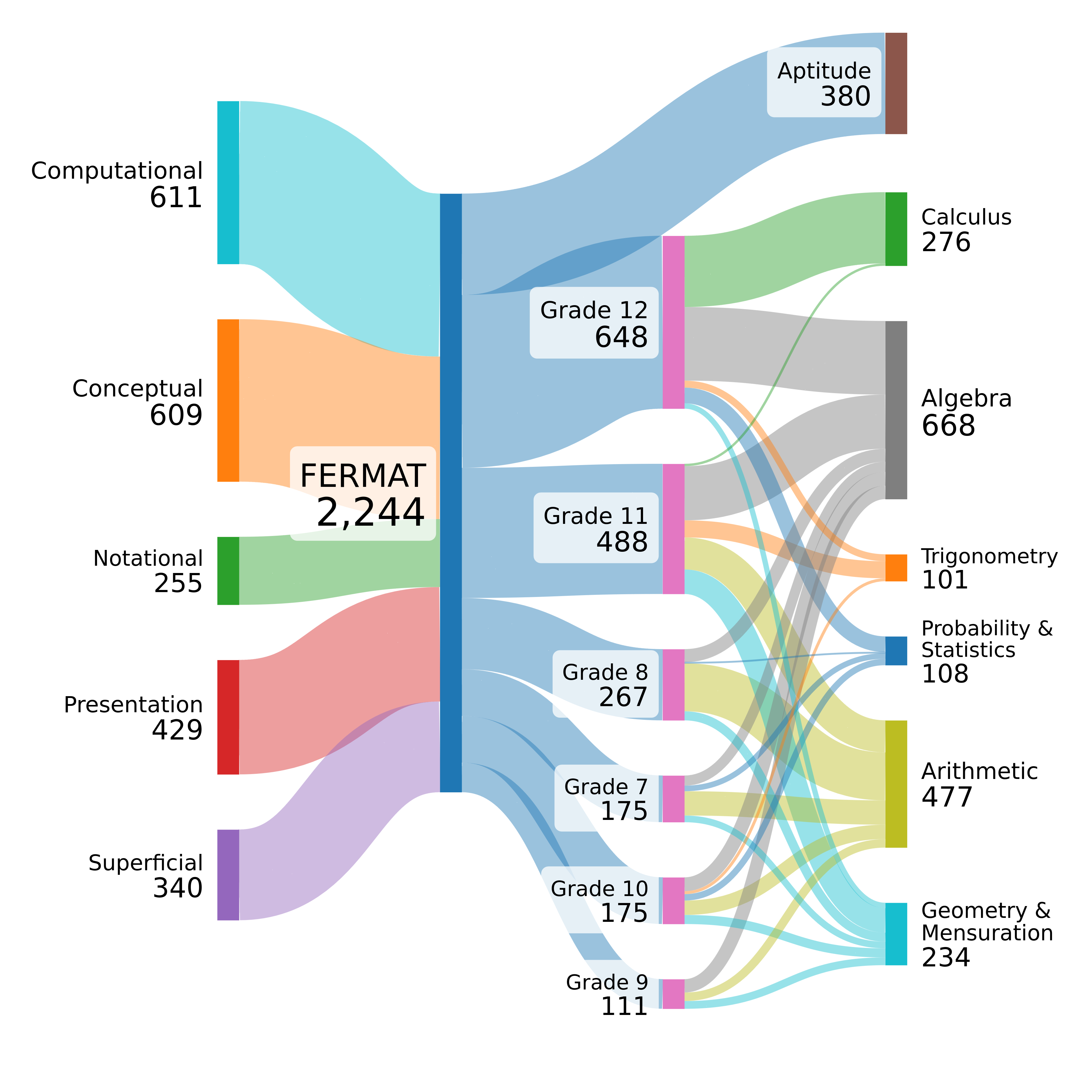}
    \caption{Distribution of different error types (left) across educational levels (middle) and math topics (right) within {\bnch}.}
    \label{fig:domain_dist}
\end{figure}

\begin{table}[h!]
\centering
\scriptsize
\begin{tabular}{l|p{0.3\columnwidth}}
\toprule
\textbf{Domain} & \textbf{Subdomains} \\
\midrule
Arithmetic & Decimals, Exponents, Factorization, Fractions, Percentages, Proportion, Ratio, Squares, Cubes, Arithmetic Progression, Permutation, Combination, Sequences \\ 
\midrule
Algebra &  Complex Numbers, Determinants, Expressions, Linear Equations, Linear Inequalities, Matrices, Polynomial, Relations, Functions,  Sets, Vectors \\
\midrule
Mensuration \& Geometry & 3D Geometry, Circles, Ellipse, Hyperbola, Lines, Parabola, Perimeter, Polygon, Surface Area, Triangles, Volume \\ 
\midrule
Calculus & Continuity, Definite Integral, Derivatives, Differential Equations, Differentiability, Indefinite Integral, Limits, Maxima Minima, Area Under Curve \\ 
\midrule
Probability \& Statistics & Bayes Theorem,  Conditional Probability, Data Handling, Independent Events \\ 
\midrule
Trigonometry & Inverse Trigonometric Equations, Trigonometric Functions \\ 
\midrule
Aptitude & Quantitative Aptitude \\ 
\bottomrule
\end{tabular}
\caption{Domains and Subdomains in \bnch}
\label{tab:domains_and_subdomains}
\end{table}

\subsection{Human Verification of Perturbations}
\label{appendix:PQCS}
\begin{figure*}[h!]
    \centering
    \includegraphics[width=0.6\textwidth]{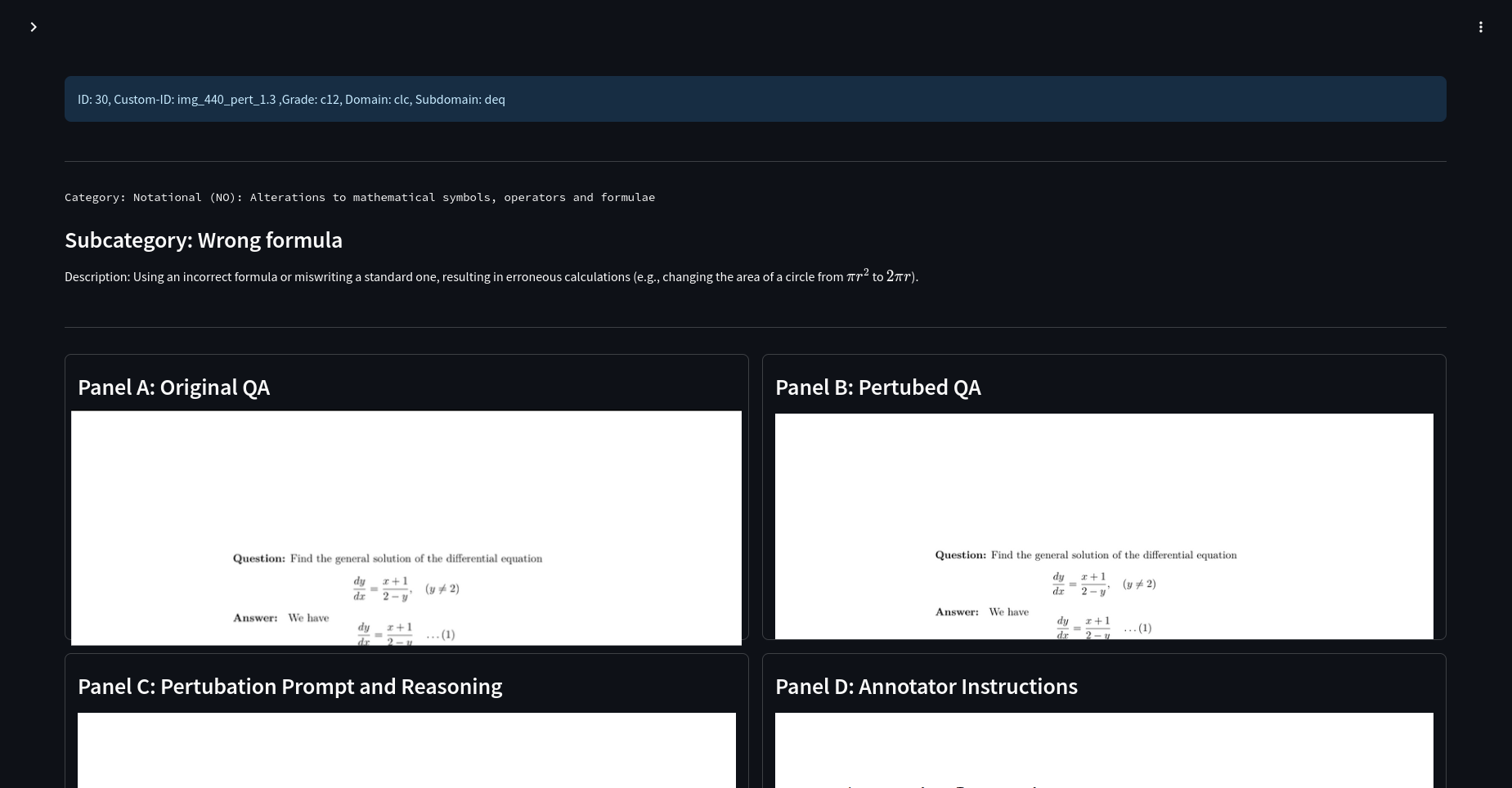}
    \caption{Interface for manual verification of perturbations (1).}
    \label{fig:PQCS_1}
\end{figure*}

\begin{figure*}[h!]
    \centering
    \includegraphics[width=0.6\textwidth]{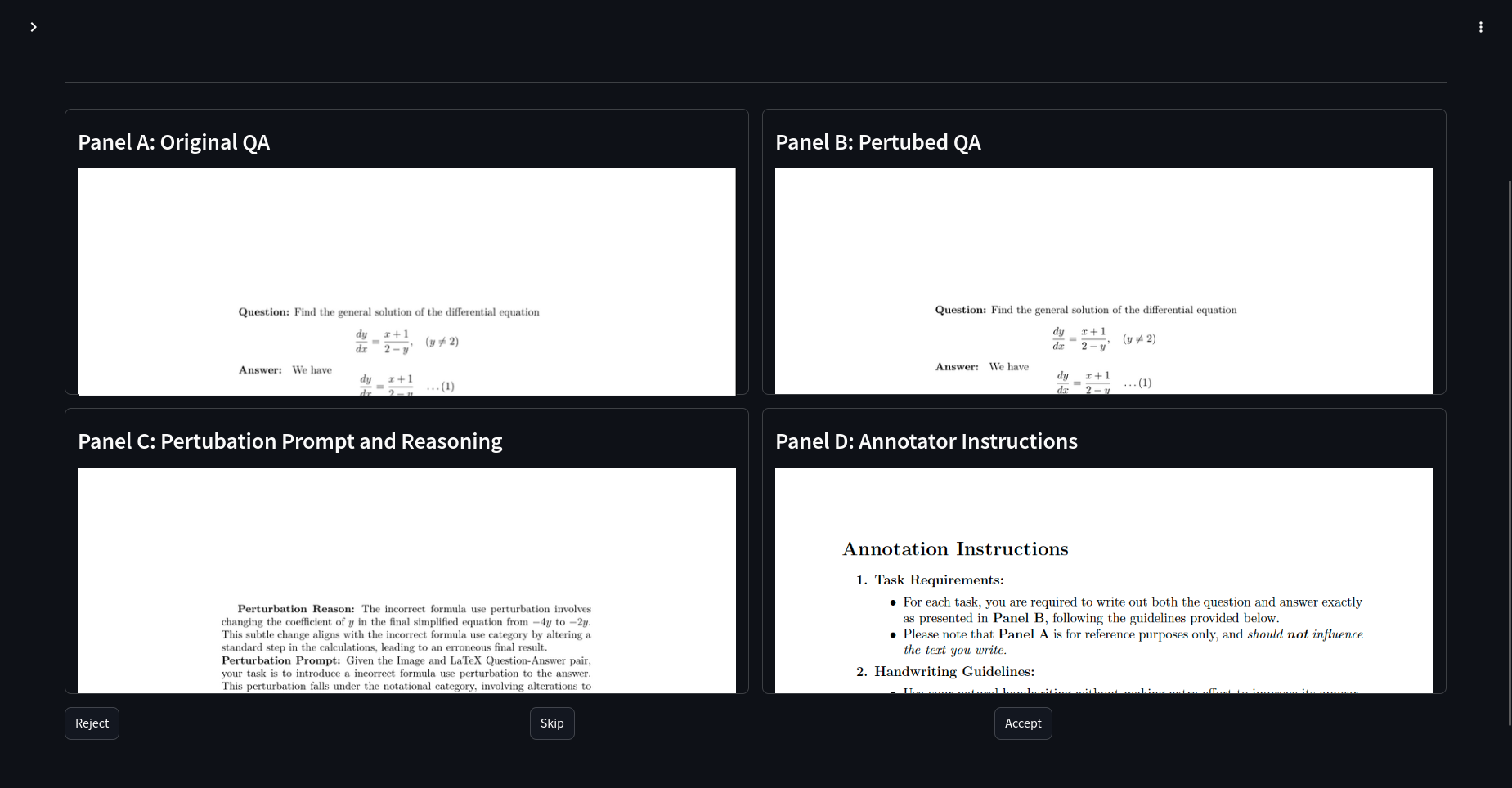}
    \caption{Interface for manual verification of perturbations (2).}
    \label{fig:PQCS_2}
\end{figure*}

We enlisted three mathematically proficient graduate students familiar with VLMs to verify the perturbations. Each annotator received task instructions, the original question-answer pair, the perturbation category, and the \gptfouro-generated perturbed pair. The annotators categorized each perturbation as either: (1) Valid Perturbation, (2) Invalid Perturbation, or (3) Not Relevant. Detailed guidelines explaining the expected perturbations and the rationale for their validity were provided.  To assist in this task, we developed a custom application, shown in Figures \ref{fig:PQCS_1} and \ref{fig:PQCS_2}. The interface enables side-by-side comparison of original and perturbed answers to facilitate accurate categorization.

Perturbations were classified as “Valid” only if they conformed to the specified perturbation category. Those irrelevant to the category or of insufficient quality were labeled as “Invalid”. Those that had minor mistakes were classified as ``Not Relevant'' and subsequently were resurrected after minor adjustments.

\subsection{Handwritten transcription}
\label{appendix:AQCS}

\begin{figure*}[h!]
    \centering
    \includegraphics[width=0.6\textwidth]{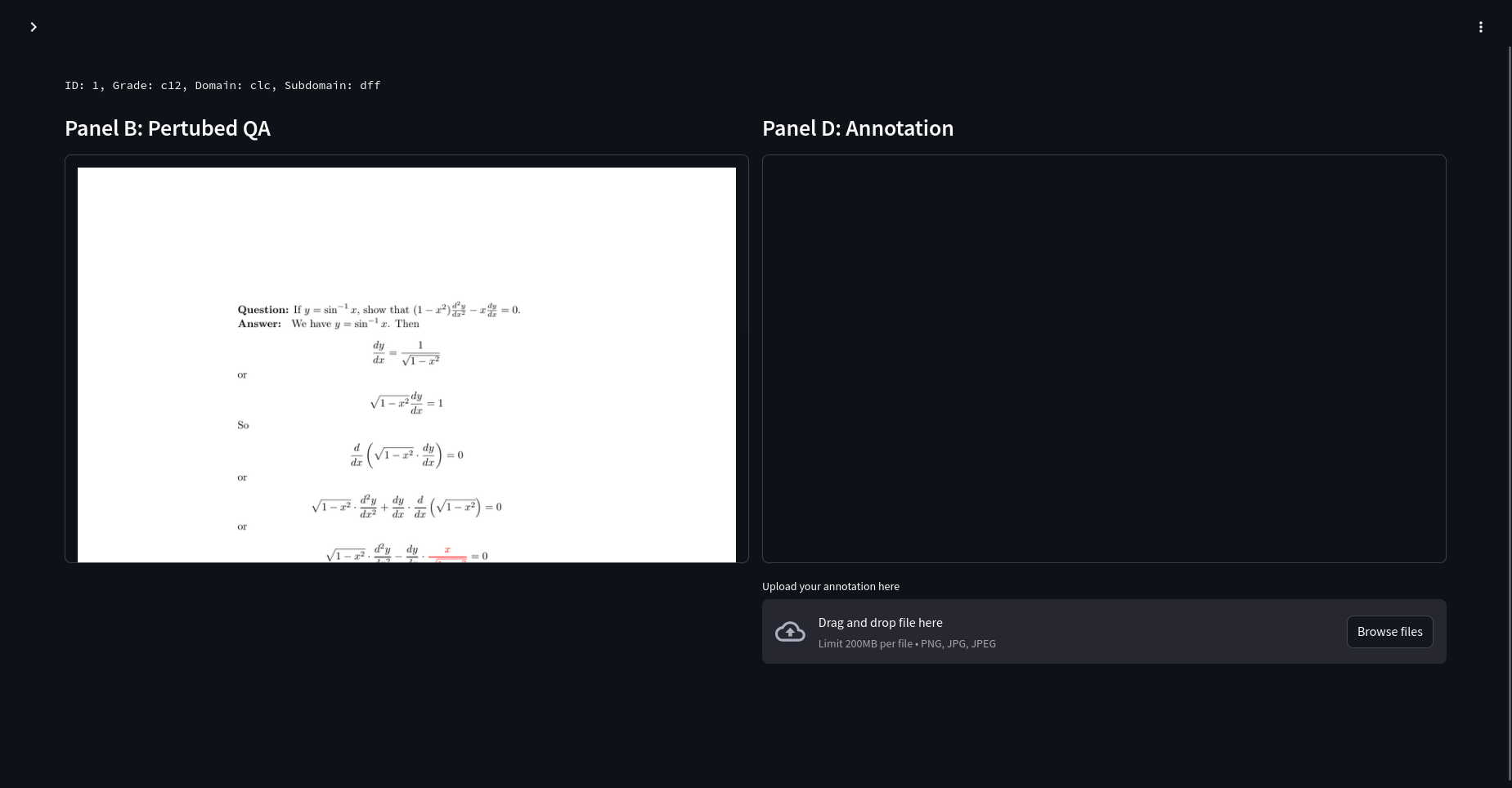}
    \caption{Interface for annotators to upload handwritten perturbed question-answer pairs (1).}
    \label{figures:Annotator_1}
\end{figure*}

\begin{figure*}[h!]
    \centering
    \includegraphics[width=0.6\textwidth]{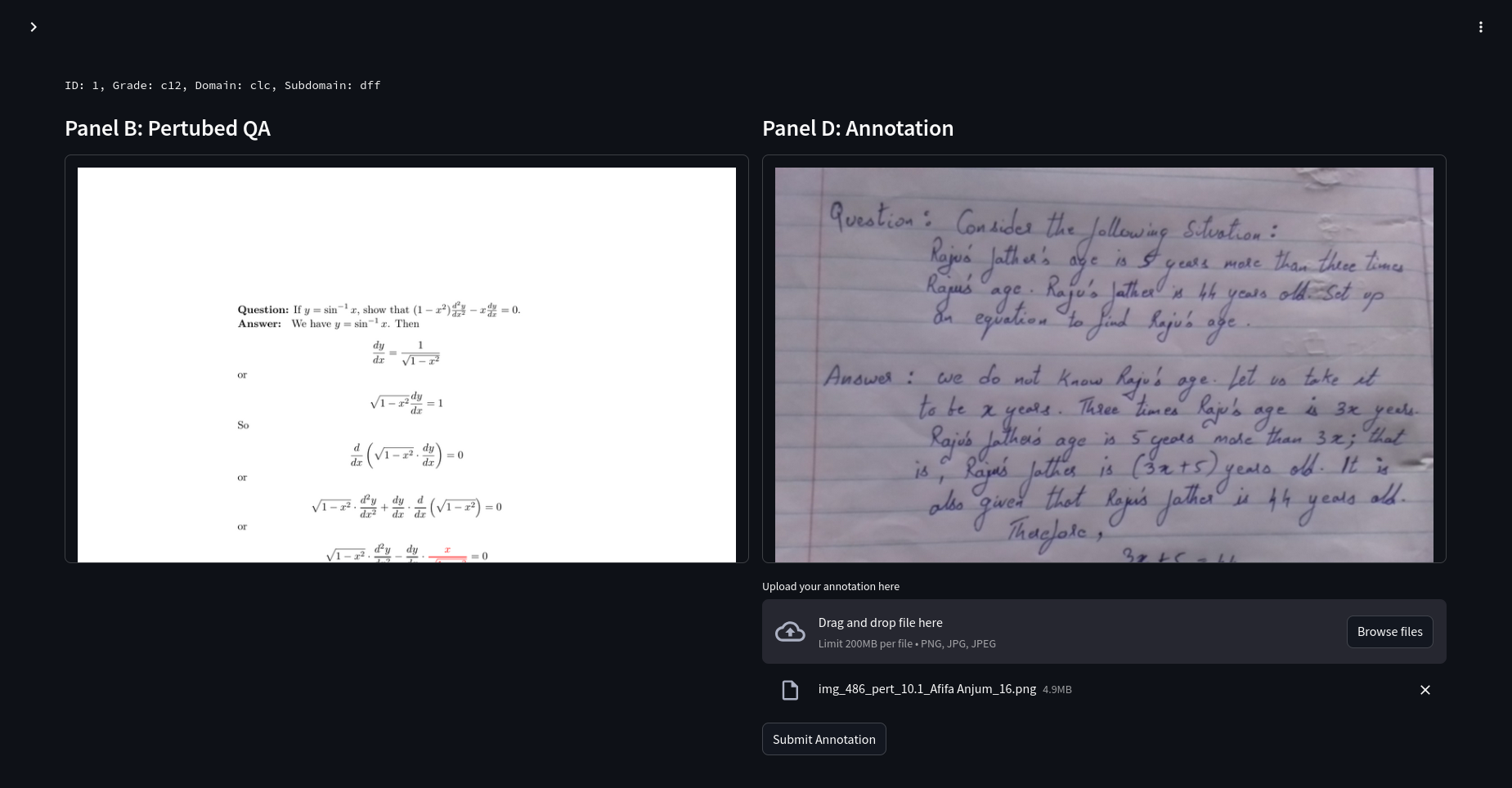}
    \caption{Interface for annotators to upload handwritten perturbed question-answer pairs (2).}
    \label{figures:Annotator_2}
\end{figure*}

We engaged 43 experienced OCR annotators to manually generate perturbed question-answer pairs, using diverse writing instruments, paper types, lighting conditions, and paper qualities. Annotators reproduced the \gptfouro~generated perturbed question-answer pairs verbatim, captured photographs, and uploaded the images directly. A dedicated application was developed to streamline this process, as illustrated in Figures \ref{figures:Annotator_1} and \ref{figures:Annotator_2}.

\section{Manual annotation quality assessment}
\label{appendix:AQCS}

\begin{figure*}[h!]
    \centering
    \includegraphics[width=0.6\textwidth]{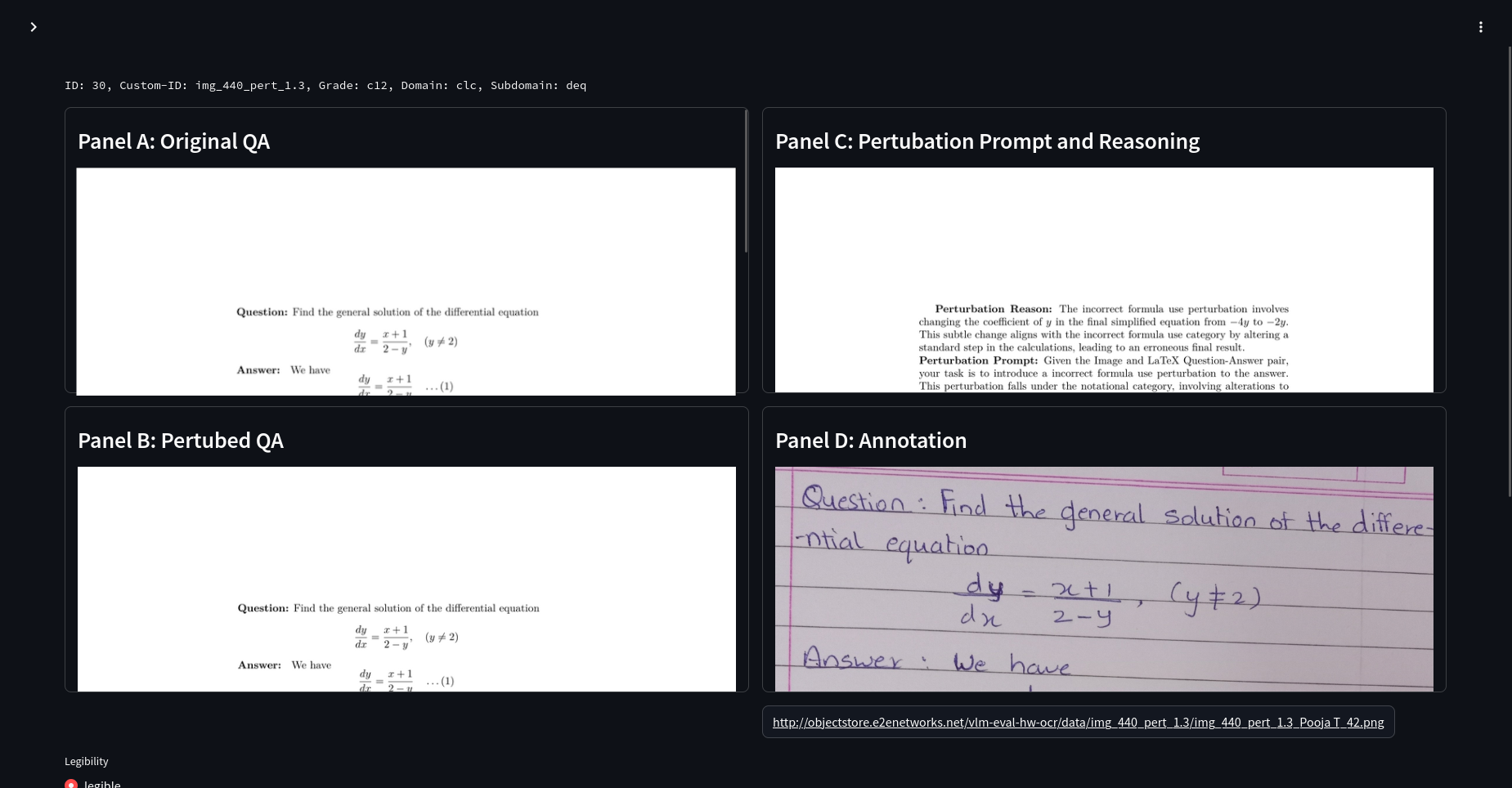}
    \caption{Interface for annotation quality assessment (1).}
    \label{figures:AQCS_1}
\end{figure*}

\begin{figure*}[h!]
    \centering
    \includegraphics[width=0.6\textwidth]{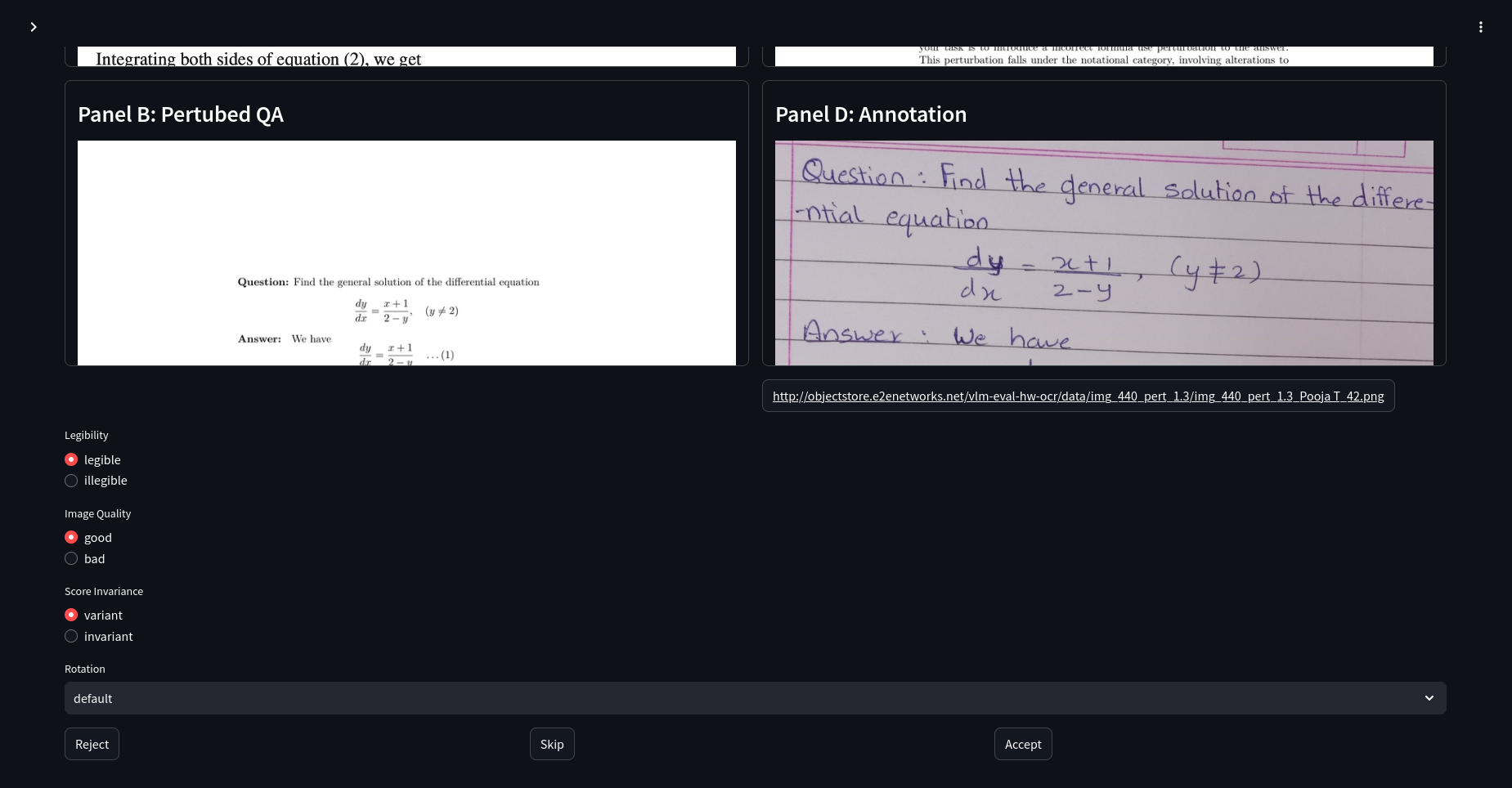}
    \caption{Interface for annotation quality assessment (2).}
    \label{figures:AQCS_2}
\end{figure*}

\begin{figure*}[h!]
    \centering
    \includegraphics[width=0.6\textwidth]{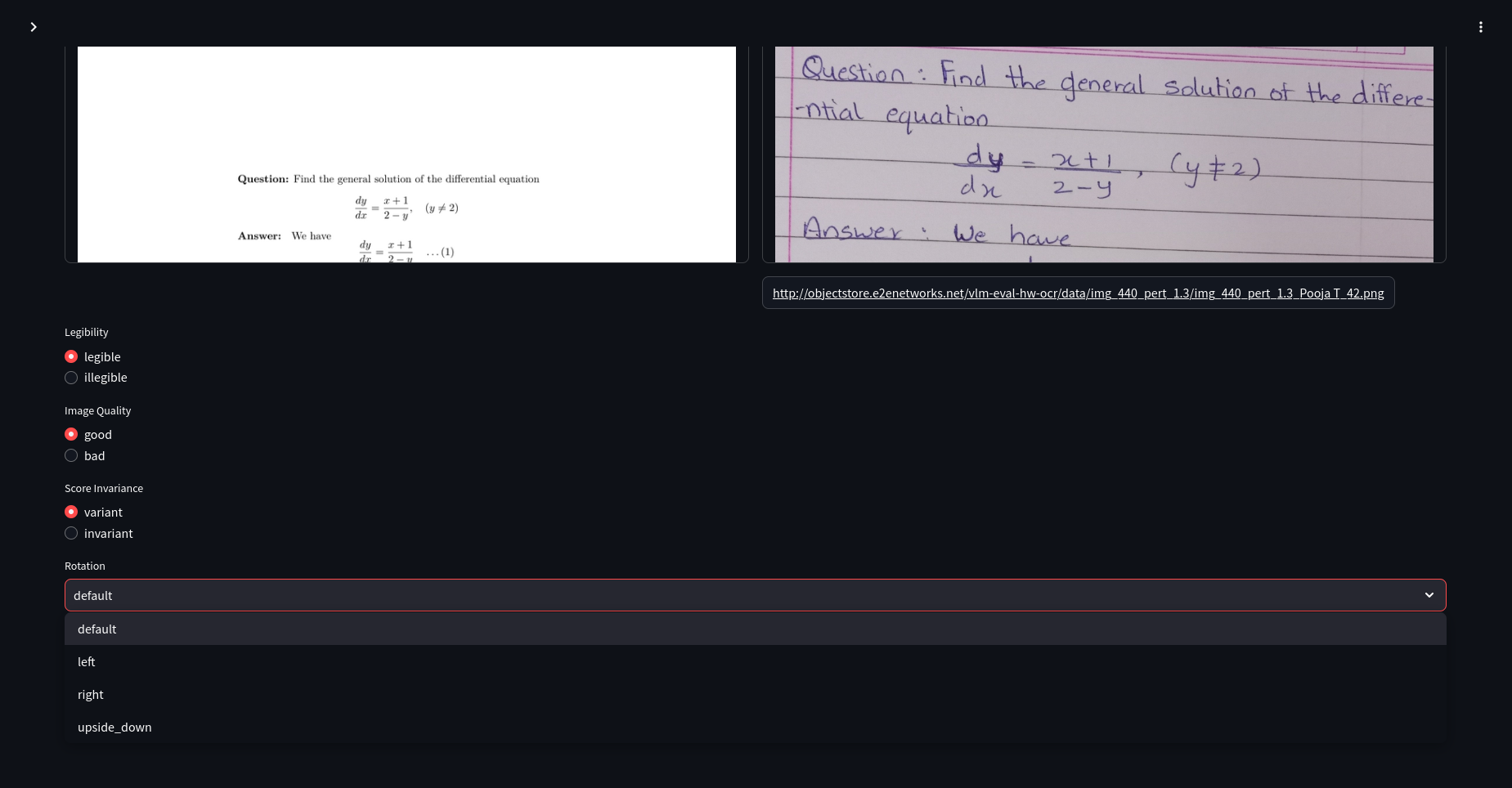}
    \caption{Interface for annotation quality assessment (3).}
    \label{figures:AQCS_3}
\end{figure*}

Three graduate students with expertise in Vision Language Models reviewed the annotations for quality assurance. Each reviewer received task instructions, the original question-answer pair, the perturbation reasoning and category, the \gptfouro-generated perturbed pair, and its handwritten version. Annotations were classified as: (1) High-Quality, (2) Low-Quality, or (3) Not Sure. The application interface used for this task is depicted in Figures \ref{figures:AQCS_1}, \ref{figures:AQCS_2}, and \ref{figures:AQCS_3}.

Each annotation was further evaluated based on the following criteria: legibility (legible or illegible), image quality (good or bad), score invariance (variant or invariant), and rotation (default, left, right, or upside down). This rigorous process ensures adherence to perturbation categories and accurate identification of score invariance, critical for benchmark quality.






\section{LLMs as Evaluators}
\label{appendix:human_vs_llm}

\subsection{Evaluator Details and Prompt Design}
Using human evaluation to assess VLM localization and correction outputs for \bnch~samples is both cost-intensive and laborious. Furthermore, this process must be repeated with the emergence of each new state-of-the-art VLM, limiting scalability and rapid adoption. To address these challenges, we employ LLMs as verifiers of VLM outputs, specifically leveraging \gptfouro. This decision is based on \gptfouro's broad adoption and strong performance on reasoning-based tasks.

\setlength{\tabcolsep}{5pt} 
\renewcommand{\arraystretch}{1.1} 
\begin{table}[h]
\centering
\small
\begin{tabular}{lr}
\toprule
\textbf{VLMs} & \textbf{LLM Accuracy} \\
\midrule
\gptfouro & 0.96 \\
\llamamini & 0.94 \\
\pixtral & 0.91 \\
\phivision & 0.94 \\
\midrule
\textbf{\textsc{Overall}} & 0.94 \\
\bottomrule
\end{tabular}
\caption{Comparison of {\gptfouro} performance with respect to human evaluation in verifying the correctness of error localization outputs across various VLMs. Higher values indicate better performance.}
\label{tab:verifier_eval}
\end{table}

We assessed the reliability of \gptfouro~as an Evaluator LLM through a controlled study involving 464 randomly selected outputs from \elnoocr~across four VLMs: \gptfouro, \llamamini, \pixtral, and \phivision. Graduate students independently evaluated the correctness of the VLMs' error localization outputs. These outputs were then provided to \gptfouro~along with detailed prompts (Figures \ref{fig:el_llm_0_prompt}, \ref{fig:el_llm_1_prompt}) outlining the scoring criteria, including explicit guidelines on awarding or withholding scores. For error localization, we prompt the LLM, denoted as $g(\cdot)$, using the VLM's output $text_{loc}$ (\S\ref{err_loc}) as the predicted text, alongside the perturbed answer ($A_{pert}$) and the explanation for the perturbation ($exp$) (\S\ref{perturbed_set_curation}) as the ground truth. The LLM is tasked with determining whether the VLM correctly localizes the error(s). This can be formally represented as $g(text_{loc}, A_{pert}, exp_{pert}) \rightarrow (reason, True/False)$. Similarly, for error correction, we prompt the LLM, $g(\cdot)$, using the VLM's corrected output ($A_{corr}$) (\S\ref{err_corr}) as the predicted solution and the original solution ($A_{gold}$) as the ground truth. The LLM is asked to verify if the VLM accurately corrected the solution. This process is represented as $g(A_{corr}, A_{gold}) \rightarrow (reason, True/False)$. Prompts are designed to cover potential output scenarios and includes comprehensive guidelines to ensure consistent scoring.

Our findings indicate that \gptfouro~achieves 94\% accuracy in aligning with human judgments of localization correctness. Table \ref{tab:verifier_eval} presents a comparison of \gptfouro's performance with human evaluation, demonstrating its effectiveness as an Evaluator LLM.

\subsection{Testing the reliability of Evaluator LLM}






We developed a dashboard to compare human and LLM performance in reasoning and decision-making. The evaluation was based on 464 randomly sampled items from the dataset, ensuring equal representation across all perturbation categories. The evaluation compared LLM reasoning with human reasoning, LLM decisions with human decisions, and LLM decisions with its own reasoning. This analysis is crucial to determine whether LLMs can effectively replace human annotators in error localization and correction tasks.

\section{VLM Performance in Error Detection}
\label{appendix:acc_fp}

\begin{table}[h]
\centering
\small
\begin{tabular}{lcccc}
\toprule
\multirow{3}{*}{\textbf{Models}} & \multicolumn{4}{c}{\textbf{Non-cascaded}}                             \\ \cline{2-5} 
                                 & \multicolumn{2}{c}{\textbf{ED}} & \multicolumn{2}{c}{\textbf{ED+OCR}} \\ \cline{2-5} 
                                 & \textbf{ACC}   & \textbf{F1}    & \textbf{ACC}     & \textbf{F1}      \\ 
\midrule
\geminiflash                 & 0.51           & 0.55           & 0.67             & 0.70             \\
\geminipro                   & 0.54           & 0.58           & 0.68             & 0.71             \\
\gptfouro                           & 0.51           & 0.55           & 0.59             & 0.63             \\
\gptfouromini                       & \textbf{0.78}  & \textbf{0.74}  & 0.73             & 0.72             \\
\llamamini                       & 0.68           & 0.68           & 0.72             & 0.70             \\
\llamabig                        & 0.66           & 0.67           & 0.70             & 0.72             \\
\pixtral                      & 0.77           & 0.72           & \textbf{0.75}    & \textbf{0.73}    \\
\pixtrallarge                     & 0.30           & 0.27           & 0.61             & 0.65             \\
\phivision                         & 0.70           & 0.70           & 0.59             & 0.62             \\ 
\bottomrule
\end{tabular}
\caption{Performance of VLMs on error detection task with accuracy (\textbf{Acc}) and \textbf{F1} scores as the evaluation metrics. Higher values indicate better performance.}
\label{tab:acc_fp}
\end{table}

\begin{figure}[!t]
    \centering
    \includegraphics[width=1.0\linewidth]{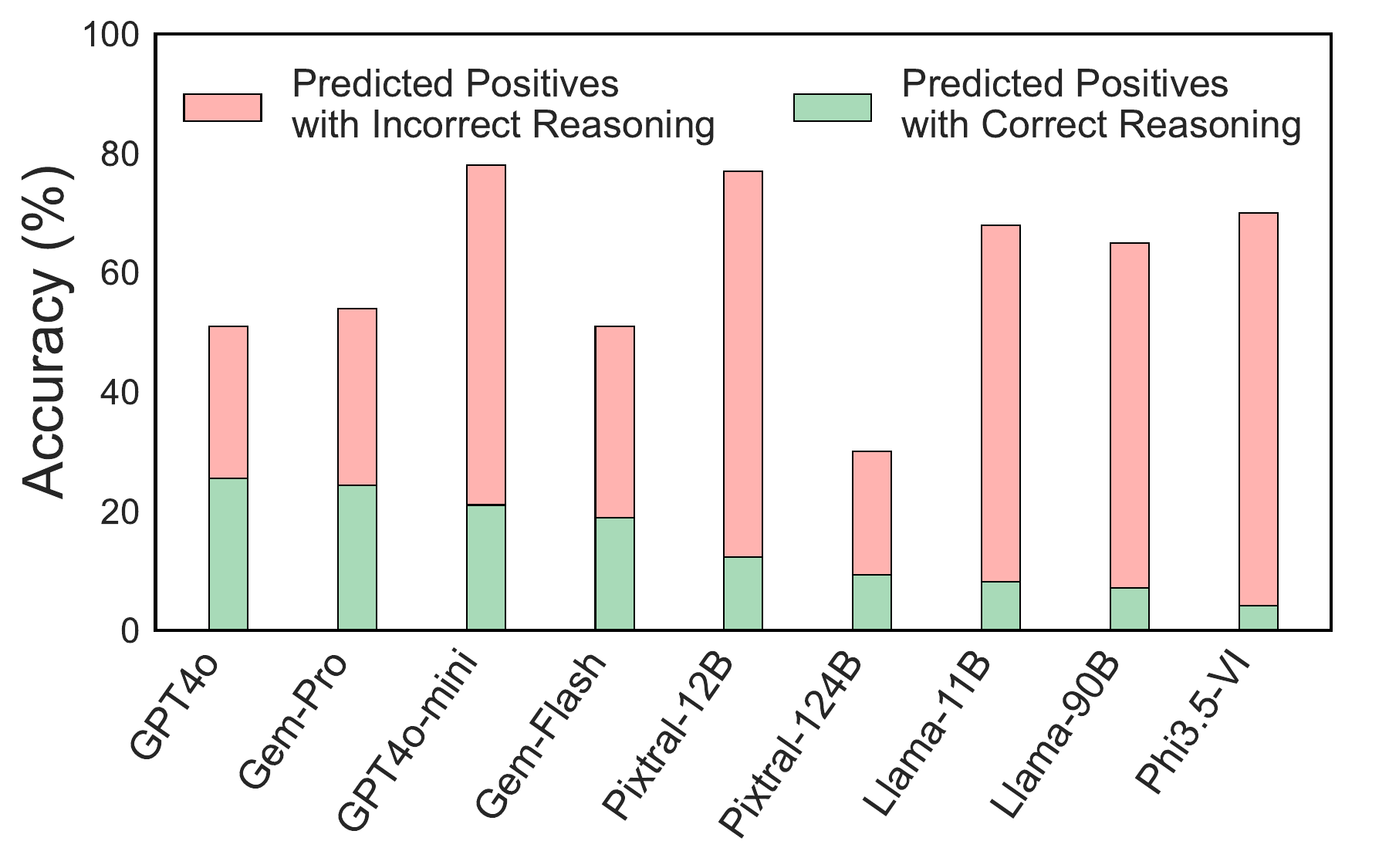}
    \caption{Performance of VLMs on the error detection task: comparing cases where predicted positives align with their reasoning against cases where they do not.}
    \label{fig:false_positives_vlms}
\end{figure}

We provide the Accuracy and F1 scores for the Error Detection task across all nine VLMs in \Cref{tab:acc_fp}. Interestingly, for Error Detection (\edcot), \gptfouro, \geminipro~and \geminiflash~models perform only slightly better than random, while \gptfouromini~outperforms all other models, a behavior that is significantly different from their performance in error localization and correction while looking at Accuracy and F1 score as a metric. To further investigate this, we analyze the explanations $(exp)$ generated as part of the \edcot~task output for all models to determine if they correctly identify errors. As shown in Figure \ref{fig:false_positives_vlms}, we observe that smaller models, including \gptfouromini, predict a high rate of positives with incorrect reasoning, indicating that these models incorrectly classify many instances as errors. Given the class imbalance in \bnch, this results in inflated Accuracy and F1 scores. In contrast, larger models such as \gptfouro~and \geminipro~produce significantly fewer False Positives. This finding aligns with previous research by \citet{llm_loc_corr}, which demonstrated that such models are generally more cautious in error detection. 

\section{Performance of VLMs in Cascaded Setup}
\label{appendix:vlm_cascaded_setup}
\begin{figure}[h!]
    \centering
    \includegraphics[width=0.7\columnwidth]{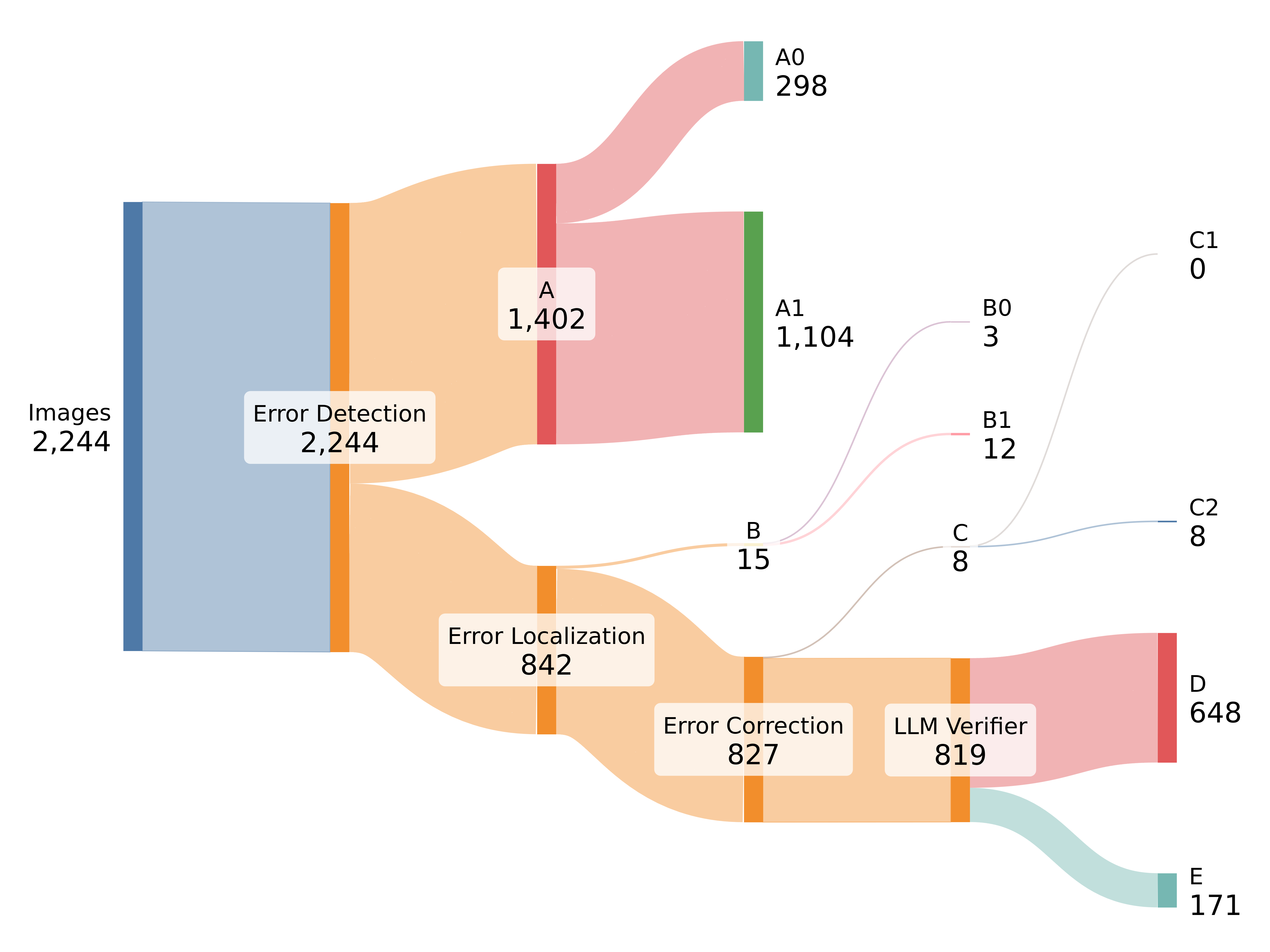}
    \caption{Breakdown of intermediate and final output proportions in GPT4o.}
    \label{figures:sankey_gpt4o}
\end{figure}

\begin{figure}[h!]
    \centering
    \includegraphics[width=0.7\columnwidth]{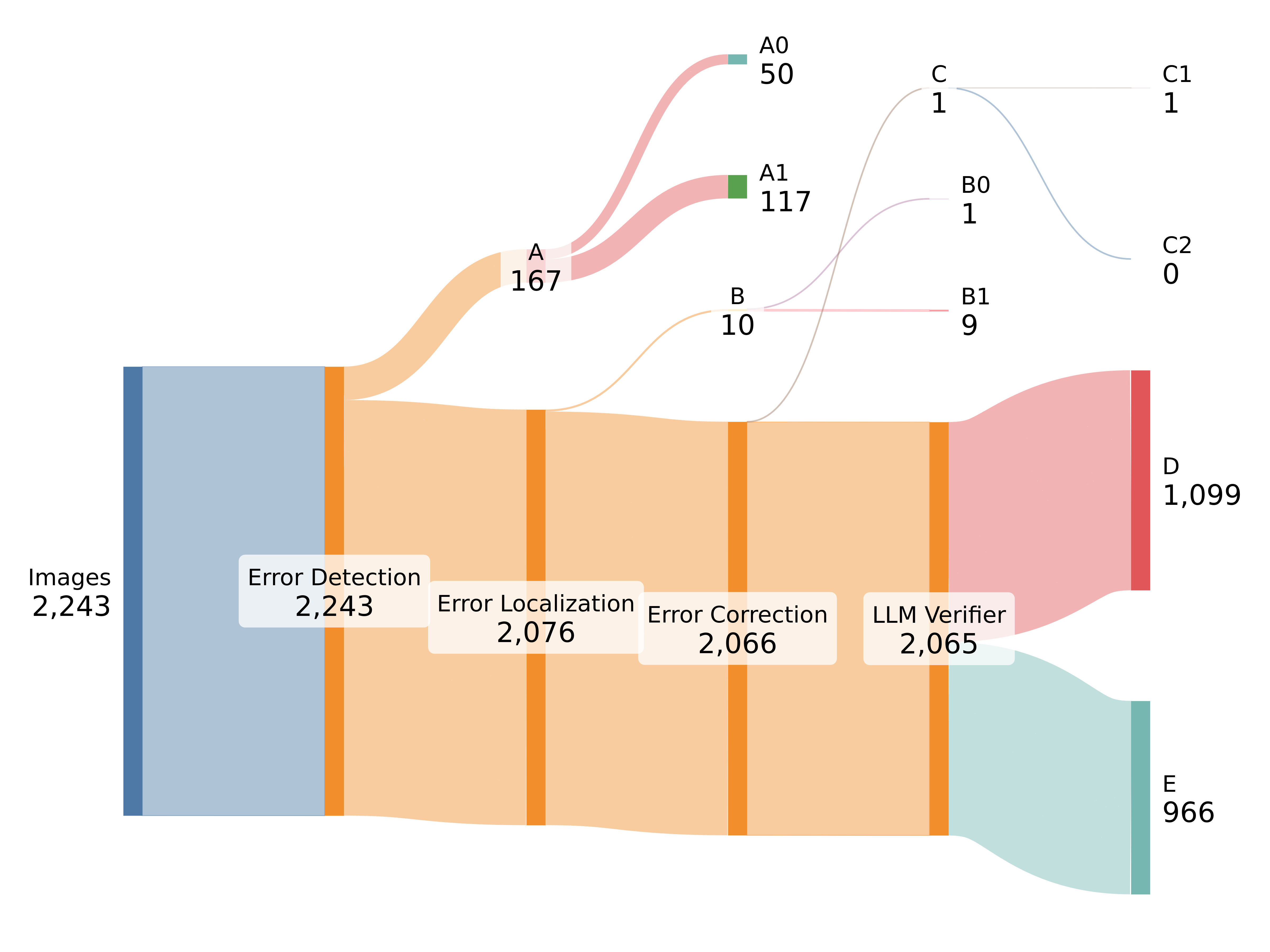}
    \caption{Breakdown of intermediate and final output proportions in GPT4o-mini.}
    \label{figures:sankey_gpt4o_mini}
\end{figure}

\begin{figure}[h!]
    \centering
    \includegraphics[width=0.7\columnwidth]{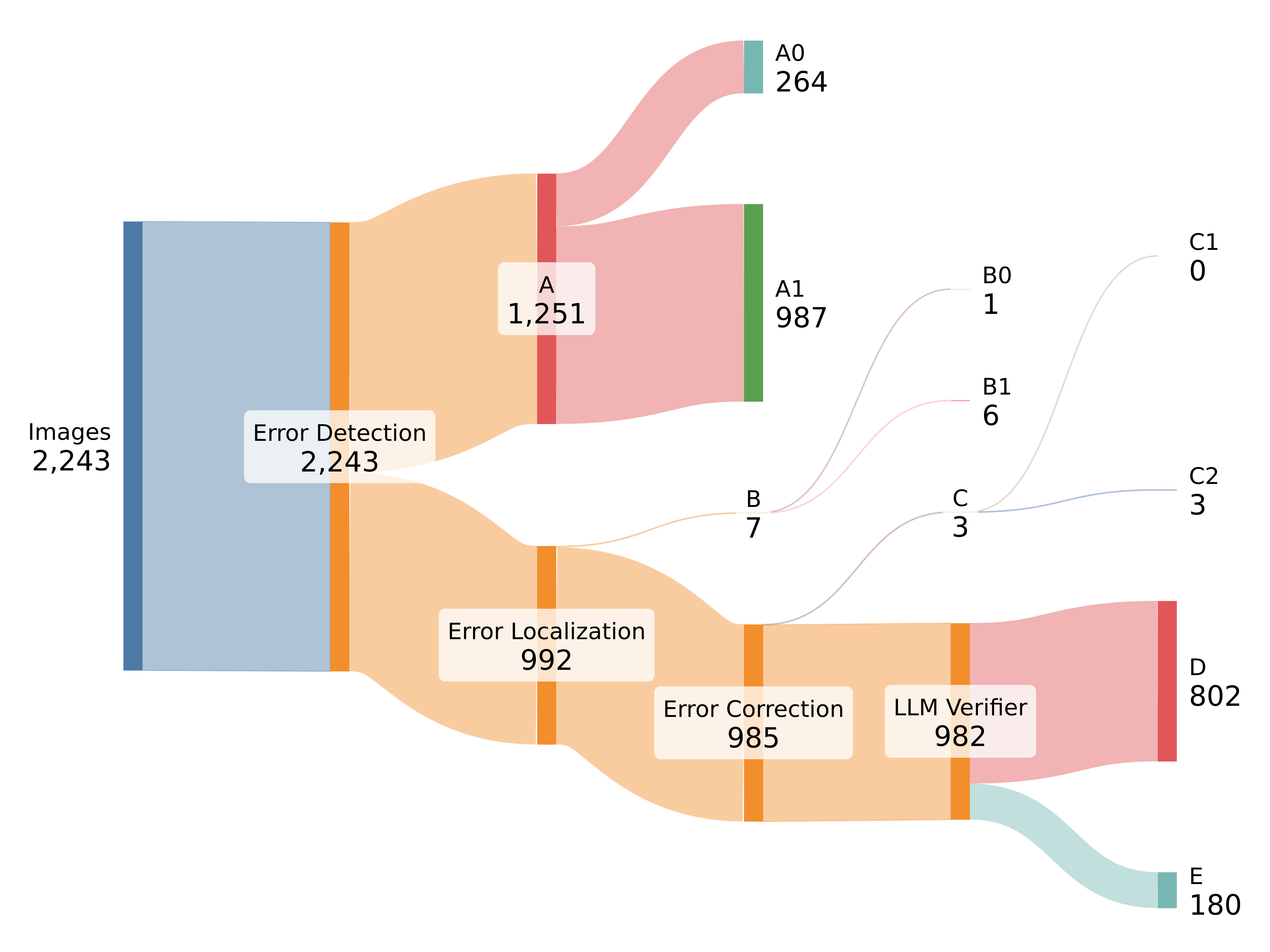}
    \caption{Breakdown of intermediate and final output proportions in Gemini Pro.}
    \label{figures:sankey_gemini_pro}
\end{figure}

\begin{figure}[h!]
    \centering
    \includegraphics[width=0.7\columnwidth]{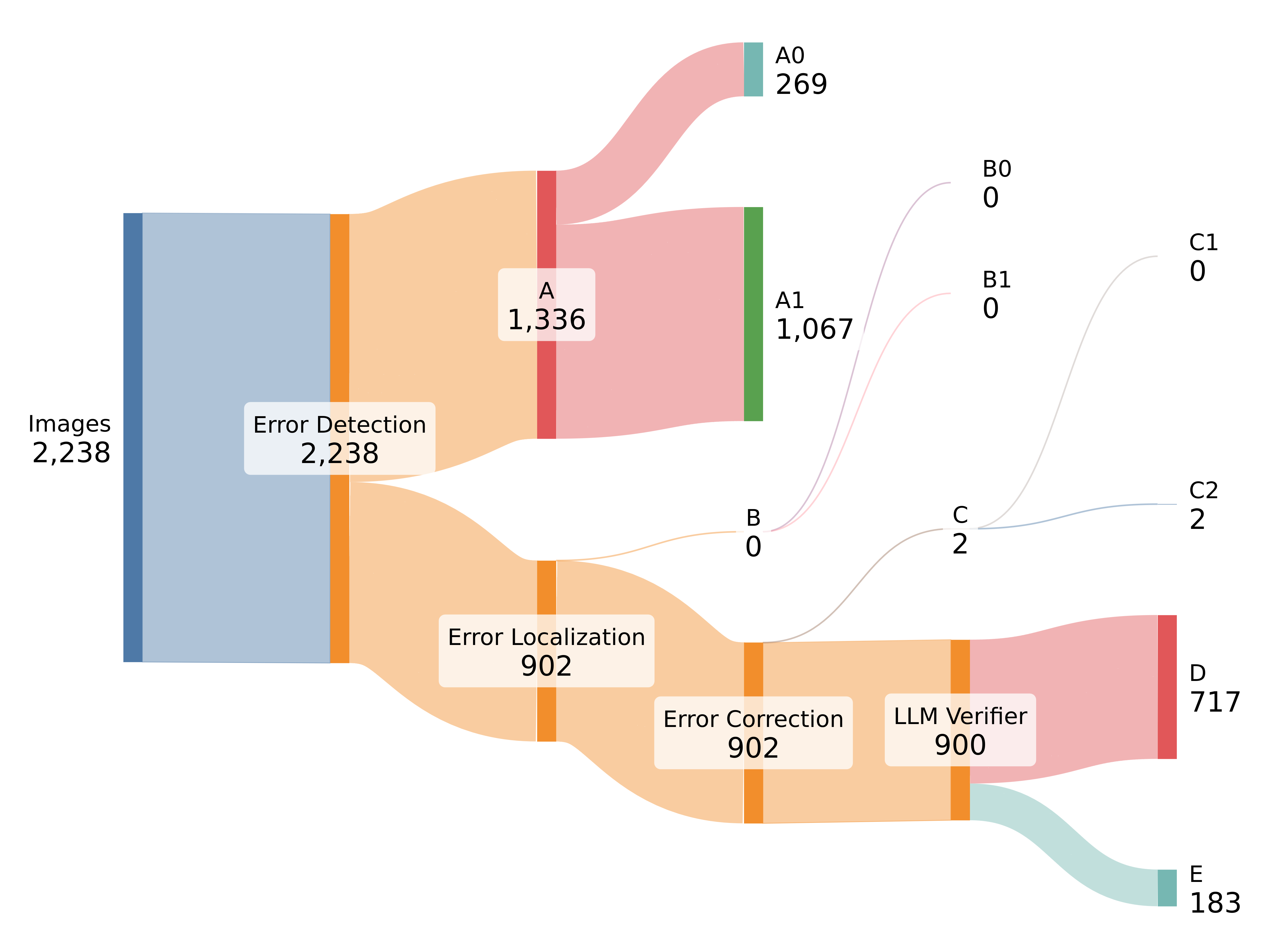}
    \caption{Breakdown of intermediate and final output proportions in Gemini Flash.}
    \label{figures:sankey_gemini_flash}
\end{figure}

\begin{figure}[h!]
    \centering
    \includegraphics[width=0.7\columnwidth]{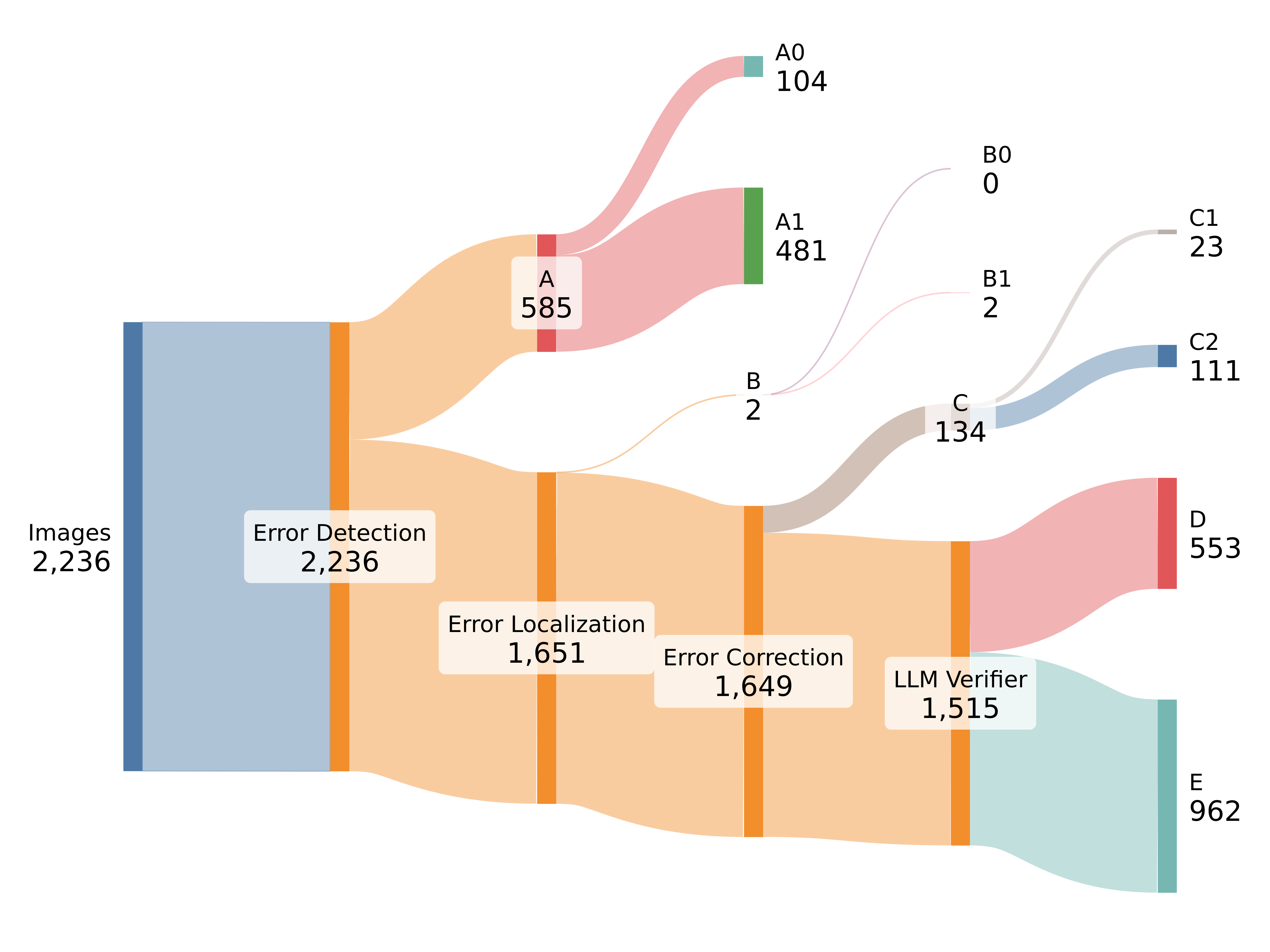}
    \caption{Breakdown of intermediate and final output proportions in LLaMA Large.}
    \label{figures:sankey_llama_large}
\end{figure}

\begin{figure}[h!]
    \centering
    \includegraphics[width=0.7\columnwidth]{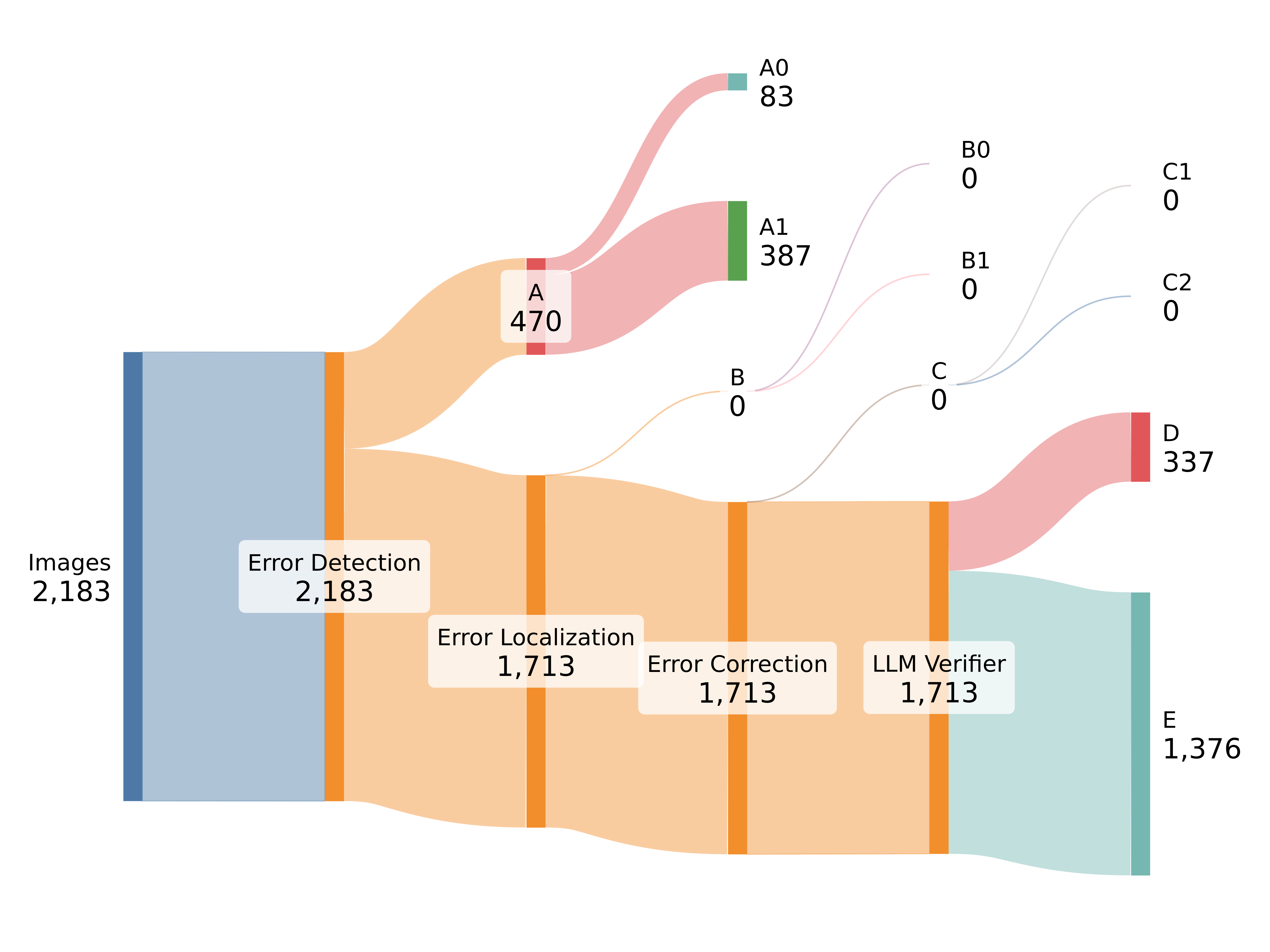}
    \caption{Breakdown of intermediate and final output proportions in LLaMA.}
    \label{figures:sankey_llama}
\end{figure}

\begin{figure}[h!]
    \centering
    \includegraphics[width=0.7\columnwidth]{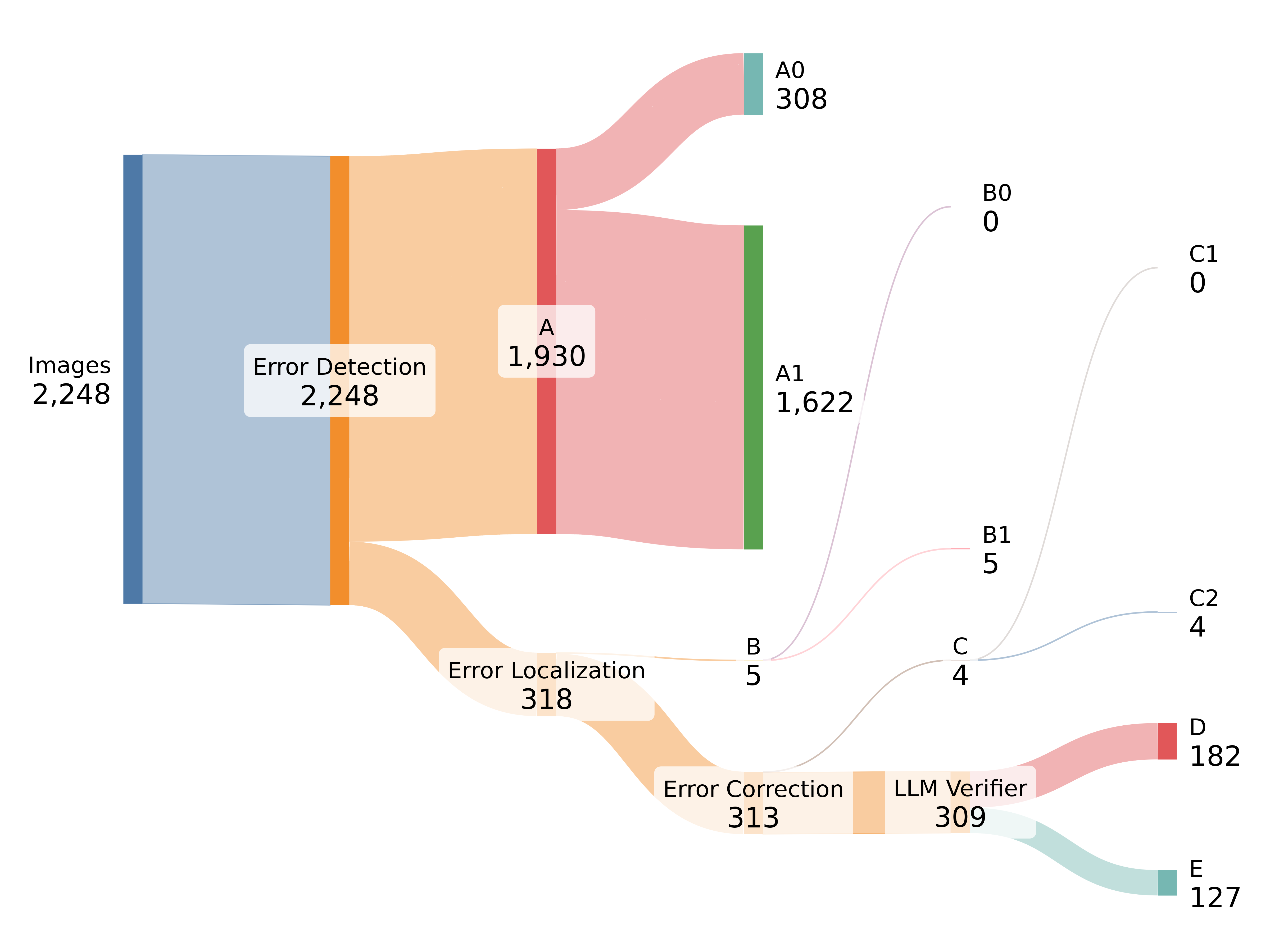}
    \caption{Breakdown of intermediate and final output proportions in Pixtral Large.}
    \label{figures:sankey_pixtral_large}
\end{figure}

\begin{figure}[h!]
    \centering
    \includegraphics[width=0.7\columnwidth]{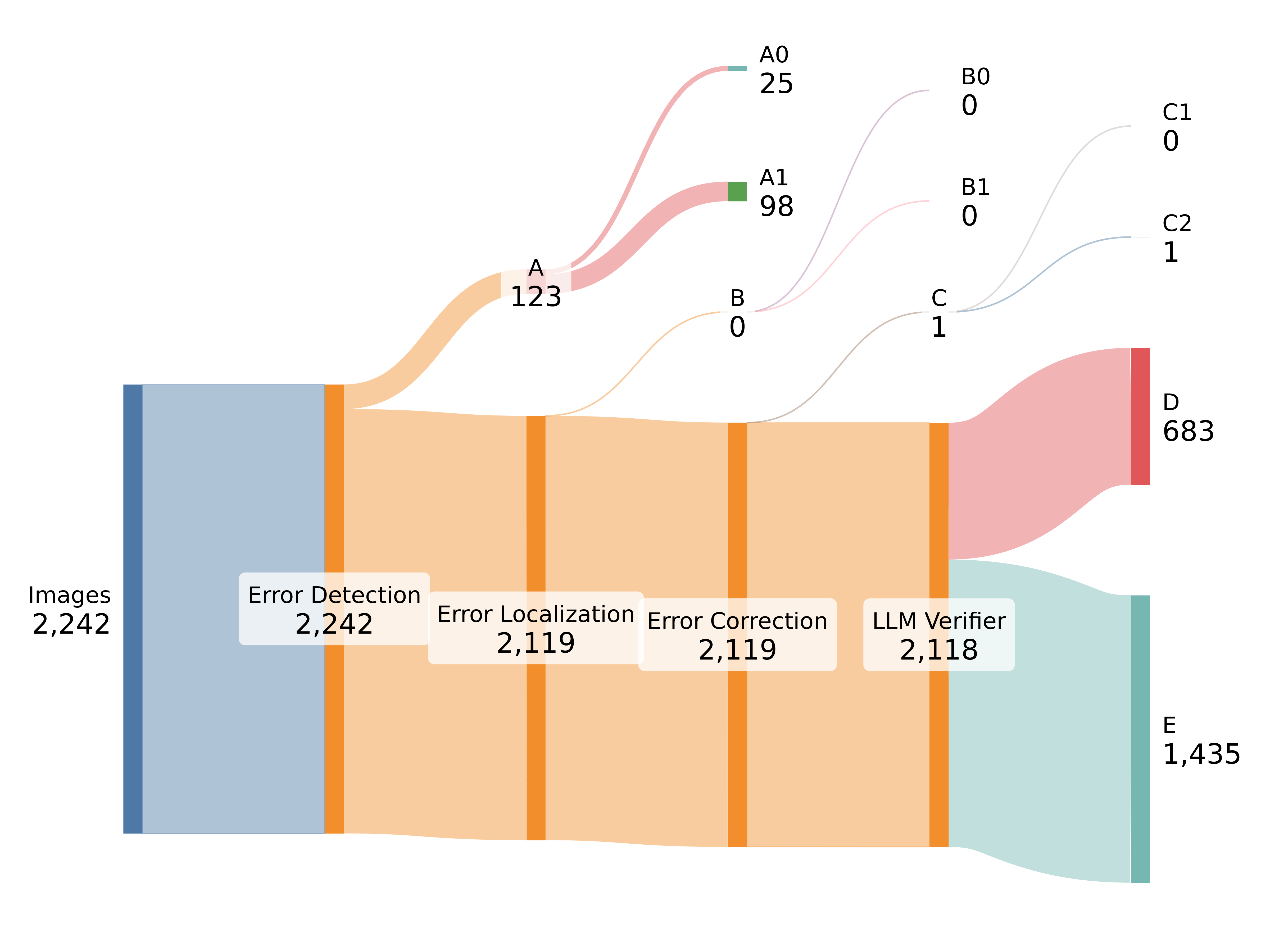}
    \caption{Breakdown of intermediate and final output proportions in Pixtral.}
    \label{figures:sankey_pixtral}
\end{figure}

\begin{figure}[h!]
    \centering
    \includegraphics[width=0.7\columnwidth]{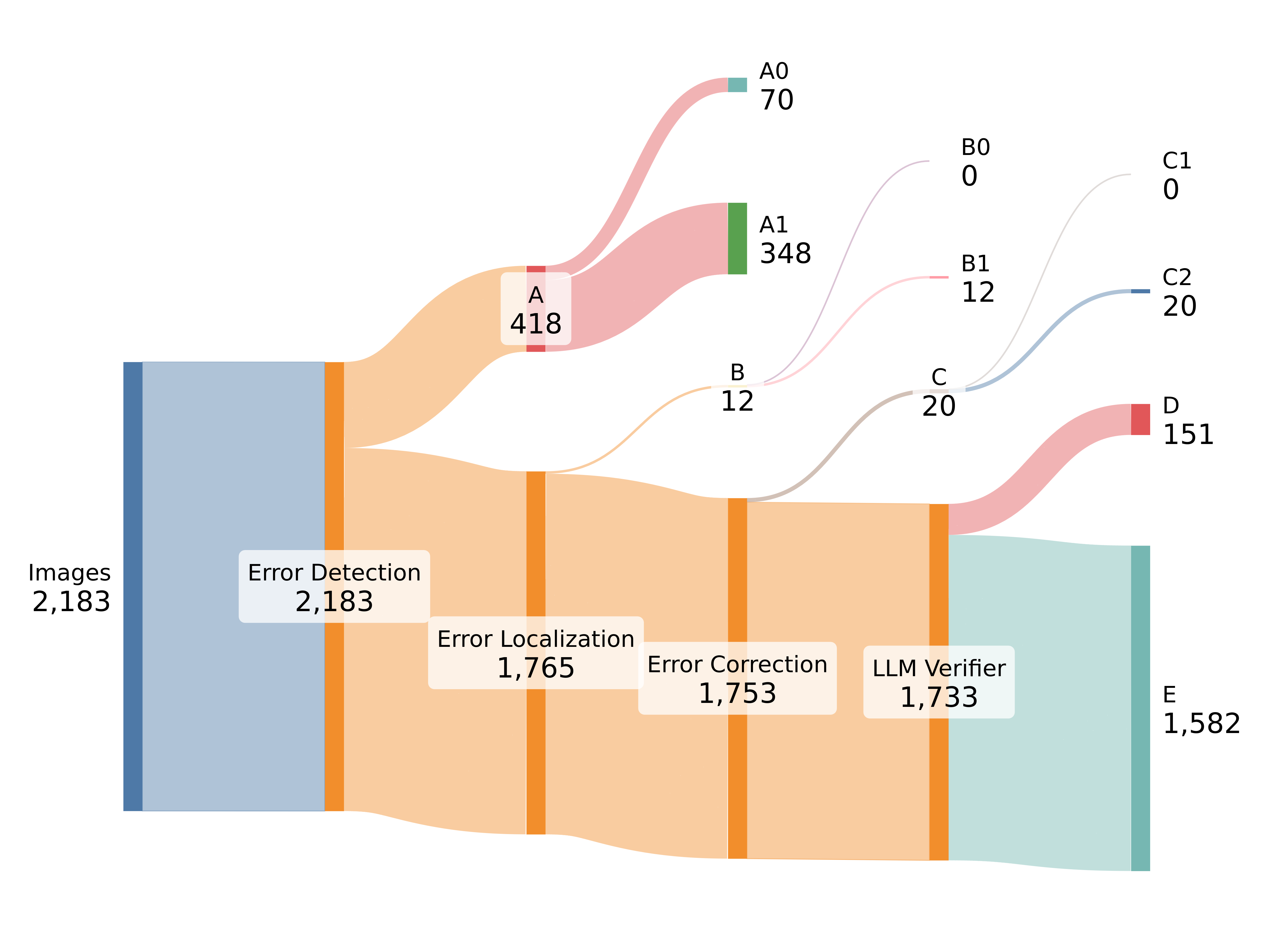}
    \caption{Breakdown of intermediate and final output proportions in Phi.}
    \label{figures:sankey_phi}
\end{figure}

We observe that bigger models like \gptfouro, \geminipro~and \geminiflash~perform worse in a cascaded Error Evaluation setup due to their cautious nature of identifying errors in a solution. On the other extreme, \pixtrallarge~gets heavily penalized due to its very high false negative prediction rate, resulting in degraded error evaluation performance. \Cref{tab:eval_non_cascaded} shows the modelwise performance on the cascaded setup. Sankey graphs illustrating the performance of VLMs in the Cascaded Setup, along with their intermediate output values, are shown in \Cref{fig:cascaded} and detailed further in \Cref{figures:sankey_gpt4o} through \Cref{figures:sankey_phi}.

\section{Prompts used for various Experiments}
\label{appendix:specific_prompts}
\begin{figure}[h!]
    \centering
    \includegraphics[width=\columnwidth]{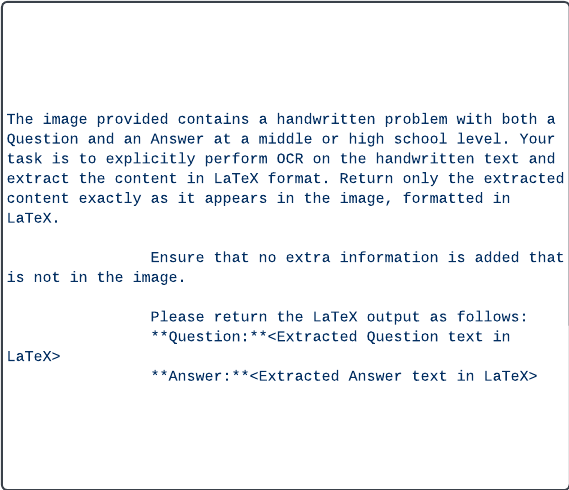}
    \caption{Prompt for OCR Extraction from Image.}
    \label{fig:ocr_prompt}
\end{figure}

\begin{figure}[h!]
    \centering
    \includegraphics[width=\columnwidth]{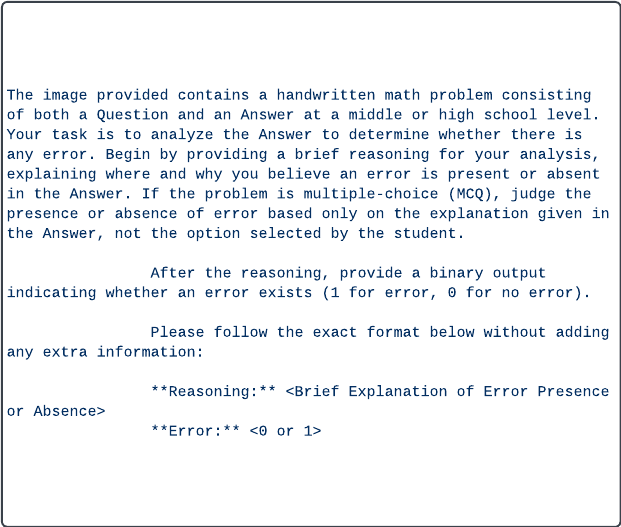}
    \caption{Prompt for Error Detection.}
    \label{fig:ed_prompt}
\end{figure}

\begin{figure}[h!]
    \centering
    \includegraphics[width=\columnwidth]{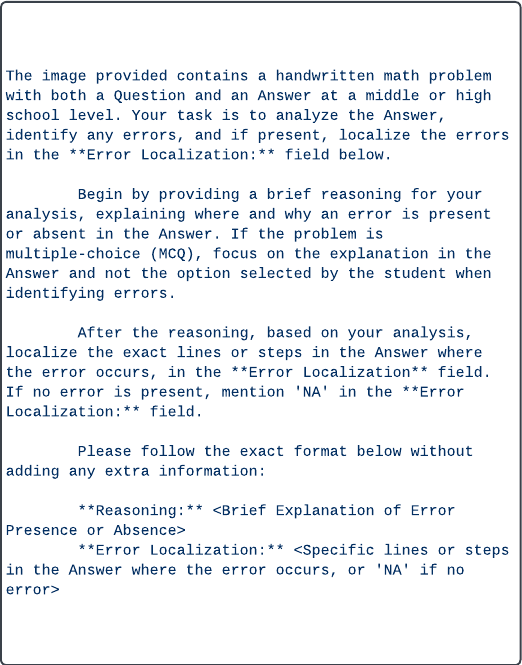}
    \caption{Prompt for Error Localization.}
    \label{fig:el_prompt}
\end{figure}

\begin{figure}[h!]
    \centering
    \includegraphics[width=\columnwidth]{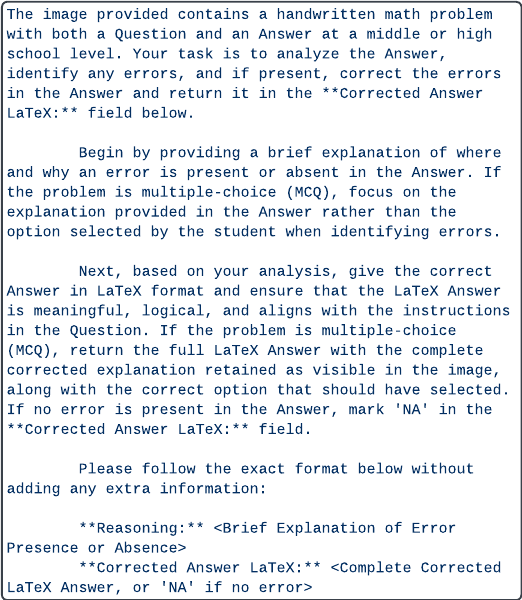}
    \caption{Prompt for Error Correction.}
    \label{fig:ec_prompt}
\end{figure}

\begin{figure}[h]
    \centering
    \includegraphics[width=0.6\textwidth]{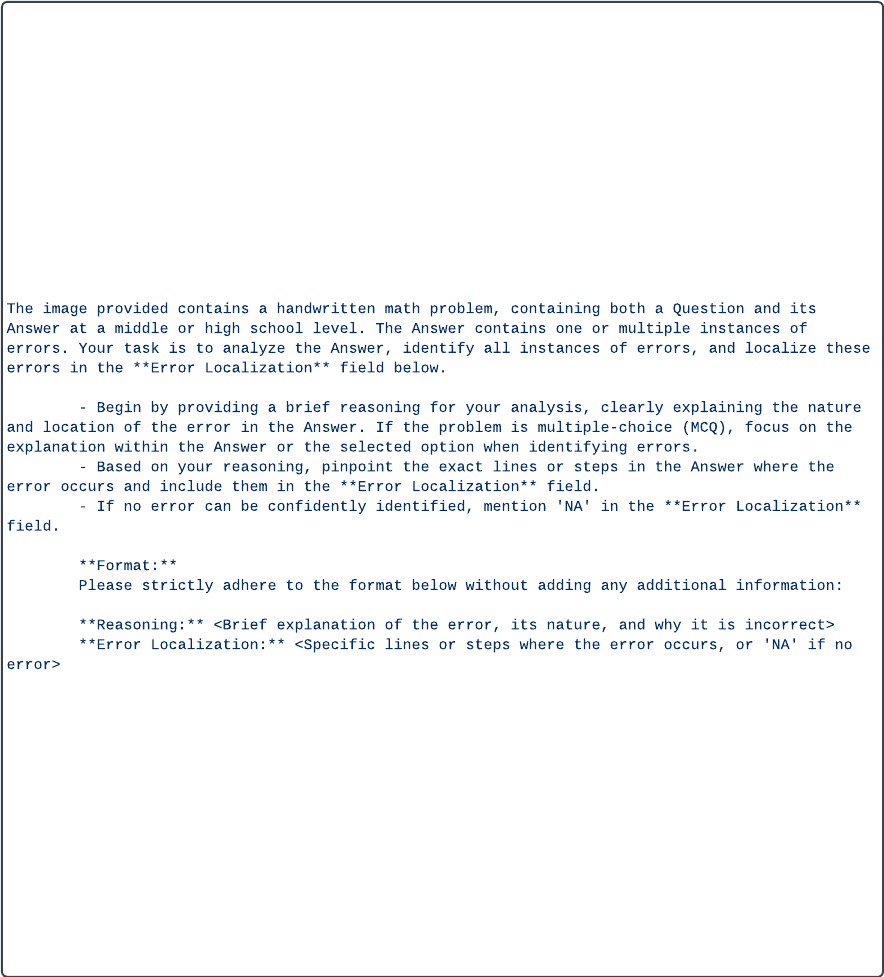}
    \caption{Prompt for Cascaded Error Localization.}
    \label{fig:c_el_prompt}
\end{figure}

\begin{figure}[h]
    \centering
    \includegraphics[width=0.6\textwidth]{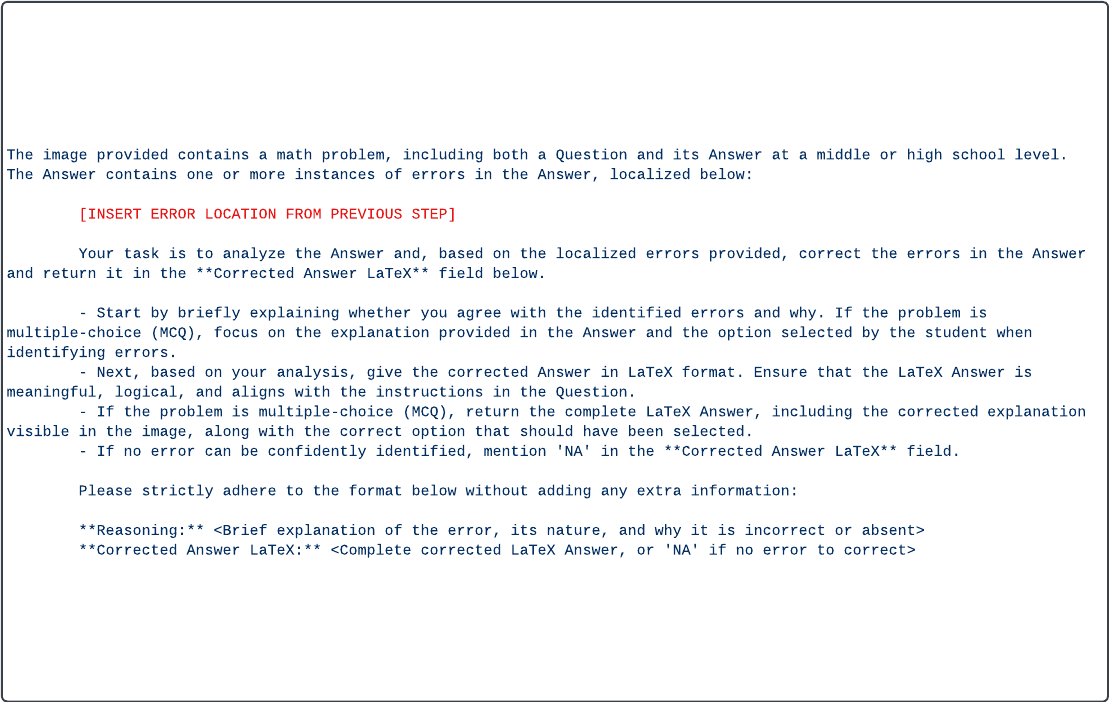}
    \caption{Prompt for Cascaded Error Correction.}
    \label{fig:c_ec_prompt}
\end{figure}

\begin{figure*}[h!]
    \centering
    \includegraphics[width=0.7\linewidth]{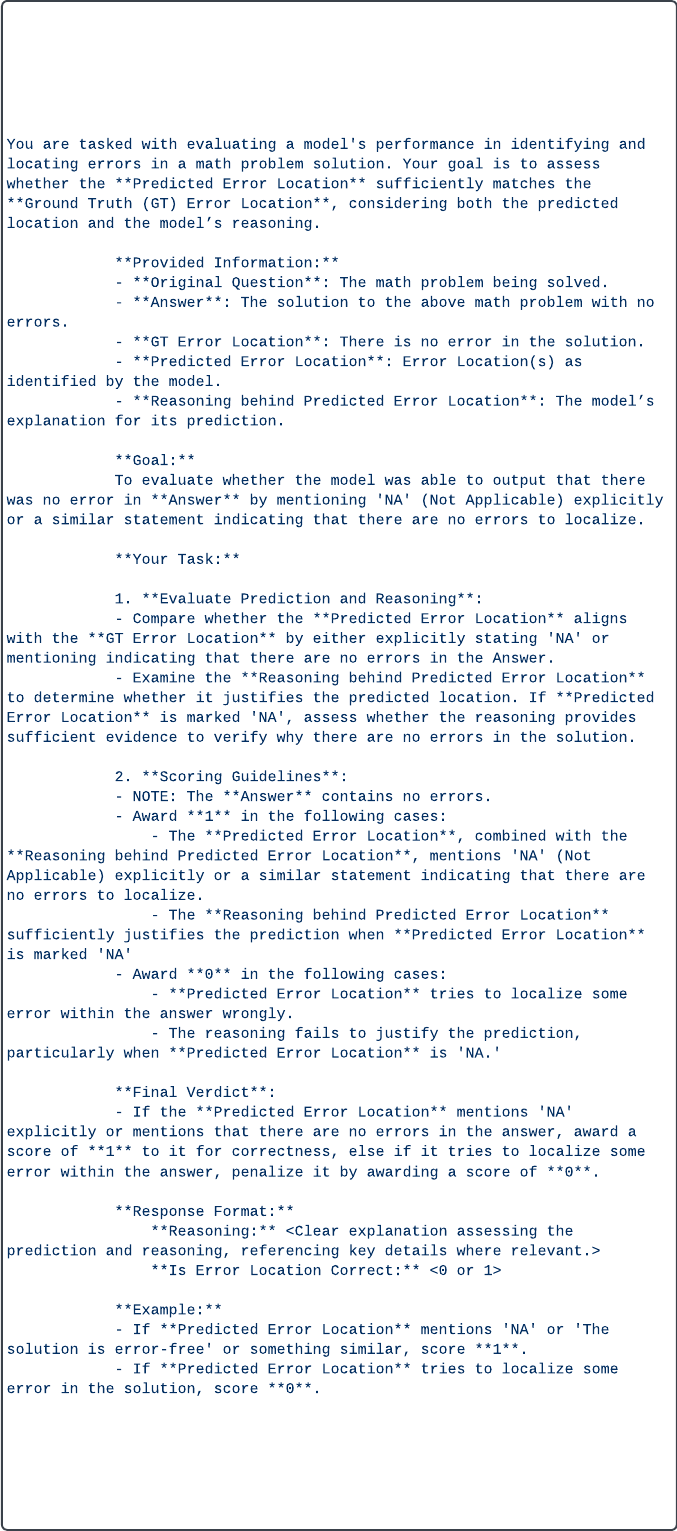}
    \caption{Prompt for LLM Verification of Error Localization outputs from VLM when the Problem Solution is error-free.}
    \label{fig:el_llm_0_prompt}
\end{figure*}

\begin{figure*}[h!]
    \centering
    \includegraphics[width=0.7\linewidth]{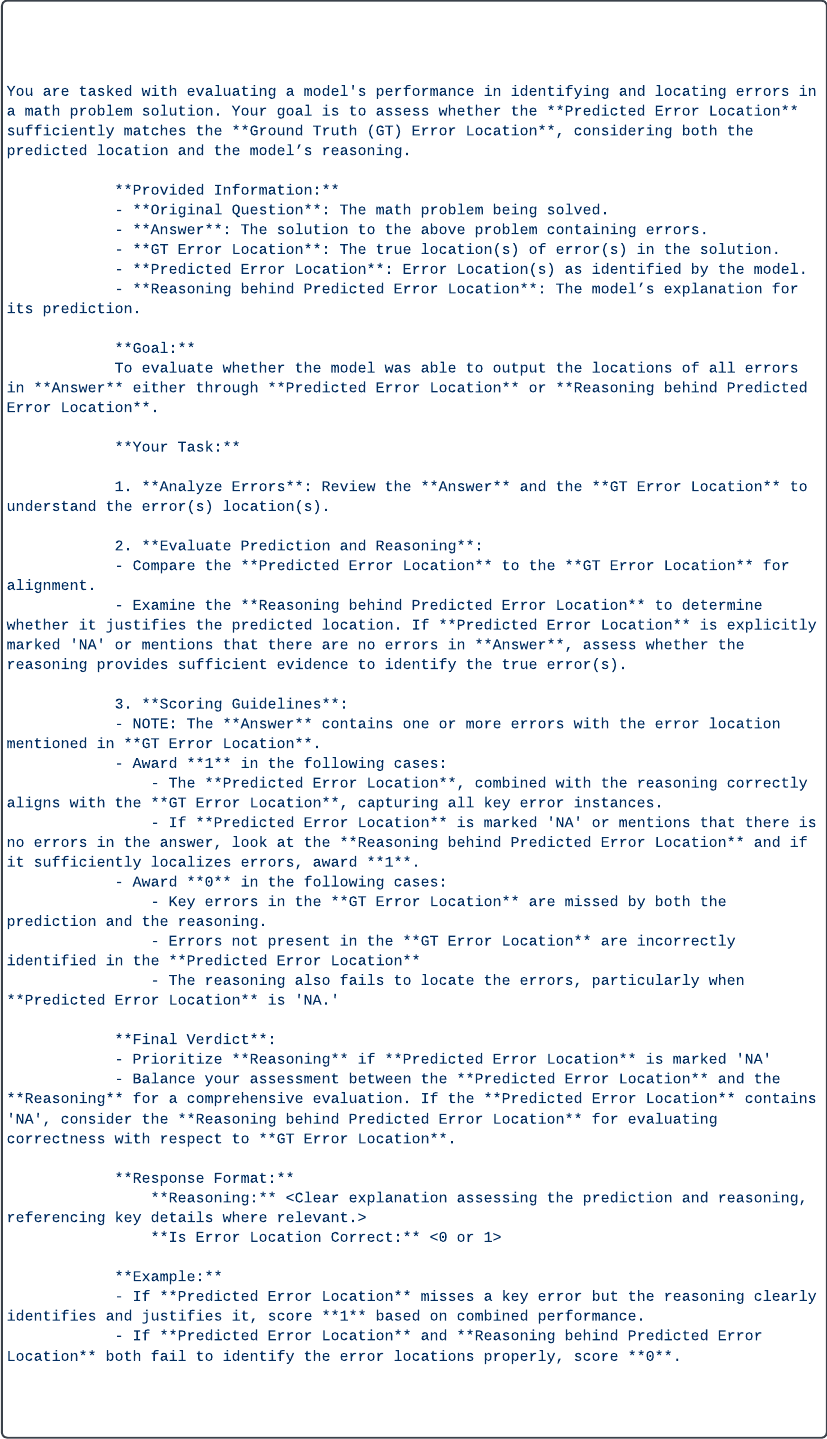}
    \caption{Prompt for LLM Verification of Error Localization outputs from VLM when the Problem Solution contains errors.}
    \label{fig:el_llm_1_prompt}
\end{figure*}

\begin{figure*}[h!]
    \centering
    \includegraphics[width=0.7\linewidth]{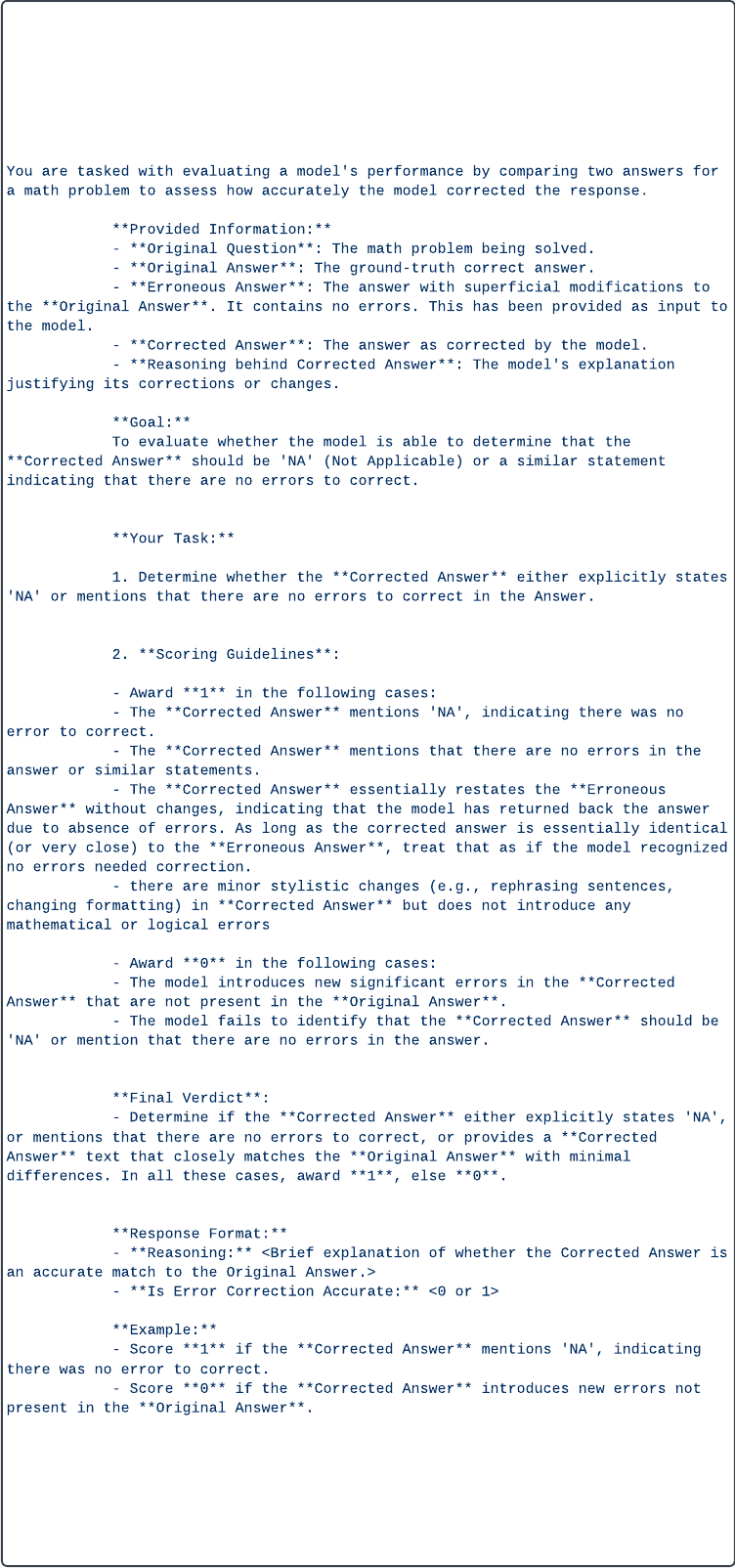}
    \caption{Prompt for LLM Verification of Error Correction outputs from VLM when the Problem Solution is error-free.}
    \label{fig:ec_llm_0_prompt}
\end{figure*}

\begin{figure*}[h!]
    \centering
    \includegraphics[width=0.7\linewidth]{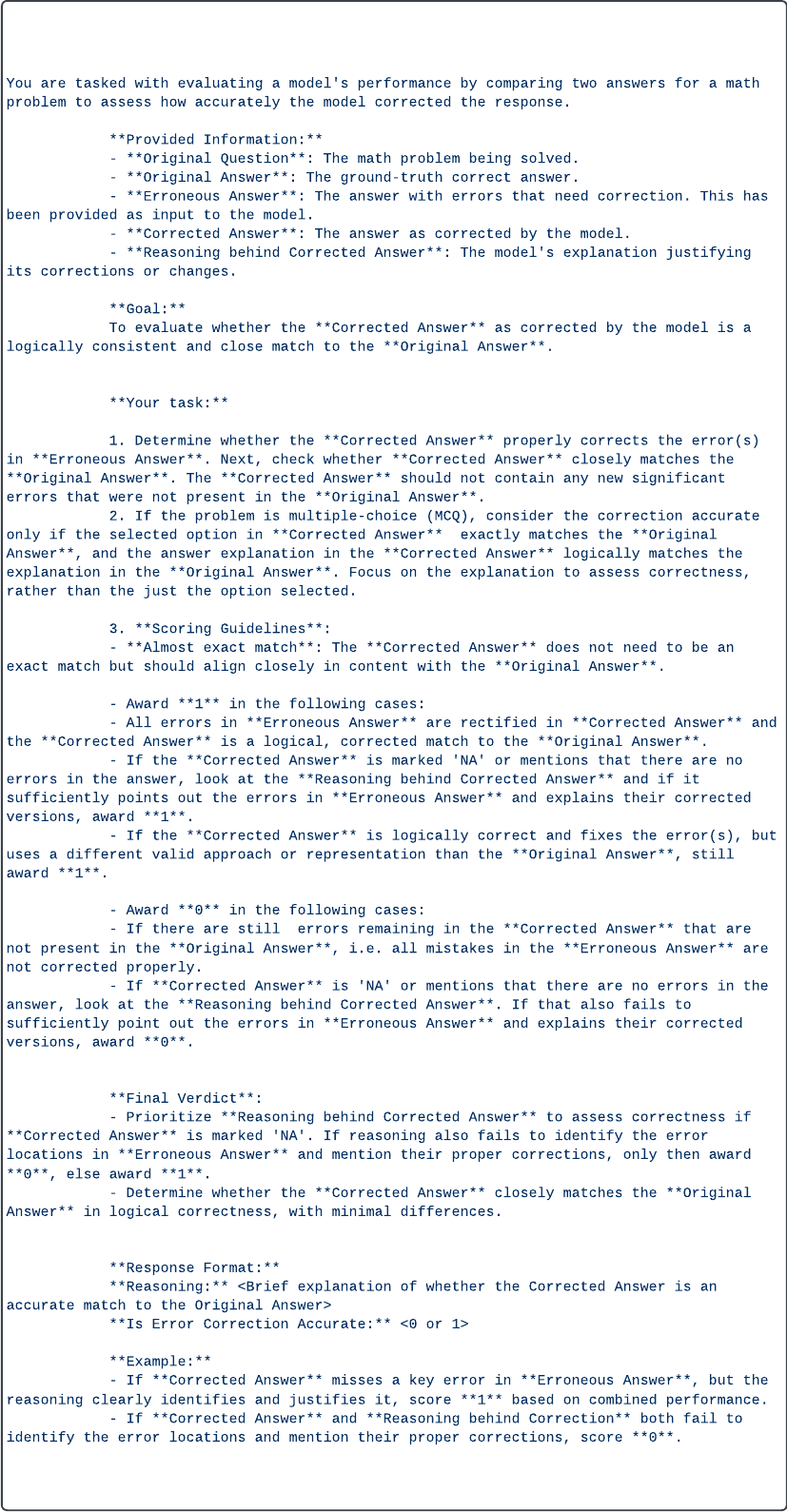}
    \caption{Prompt for LLM Verification of Error Correction outputs from VLM when the Problem Solution contains errors.}
    \label{fig:ec_llm_1_prompt}
\end{figure*}

The task-specific evaluation prompts for all Vision-Language Models (VLMs) assessed on \bnch~are detailed below in \Cref{fig:ocr_prompt} through \Cref{fig:ec_llm_1_prompt}. For each task, we ensured consistent evaluation by using identical prompts across all models and setting the sampling temperature to zero to ensure reproducibility. Similarly, for the Evaluator LLM, we employed \gptfouro~with a temperature of zero.

\begin{figure*}[h!]
    \centering
    \includegraphics[width=0.7\linewidth]{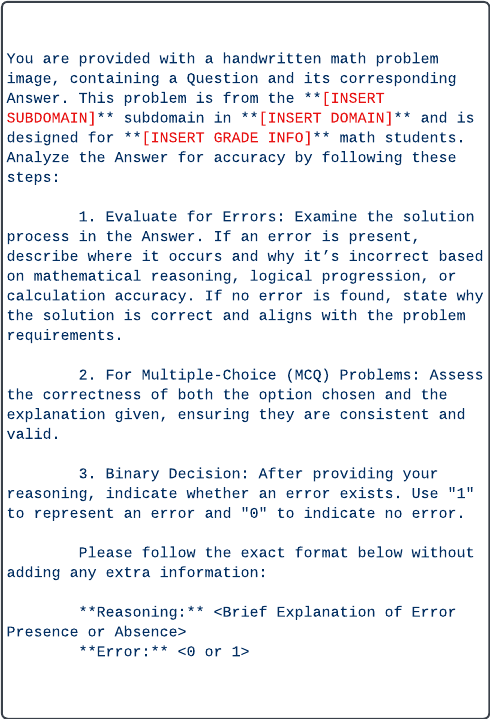}
    \caption{Level 1 Error Detection Prompt for \Cref{levels_expt}.}
    \label{fig:L1_prompt}
\end{figure*}

\begin{figure*}[h!]
    \centering
    \includegraphics[width=0.7\linewidth]{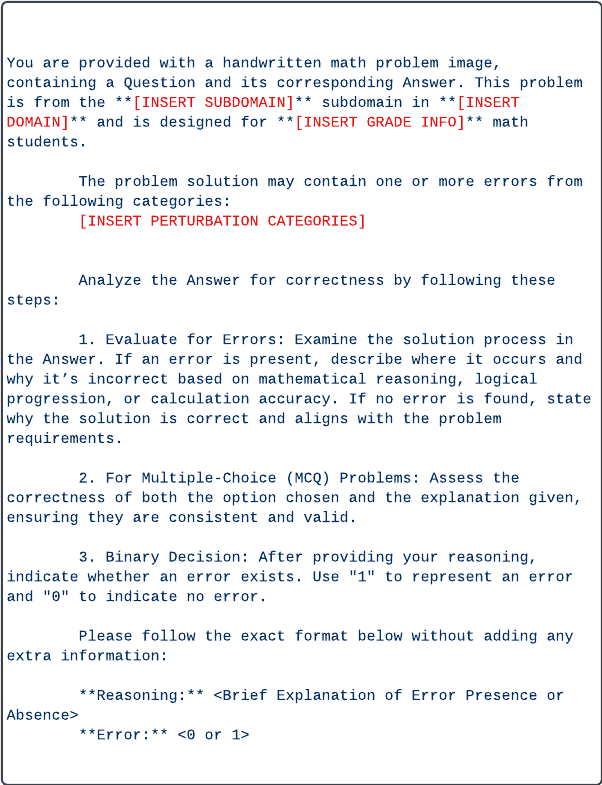}
    \caption{Level 2 Error Detection Prompt for \Cref{levels_expt}.}
    \label{fig:L2_prompt}
\end{figure*}

\begin{figure*}[h!]
    \centering
    \includegraphics[width=0.7\linewidth]{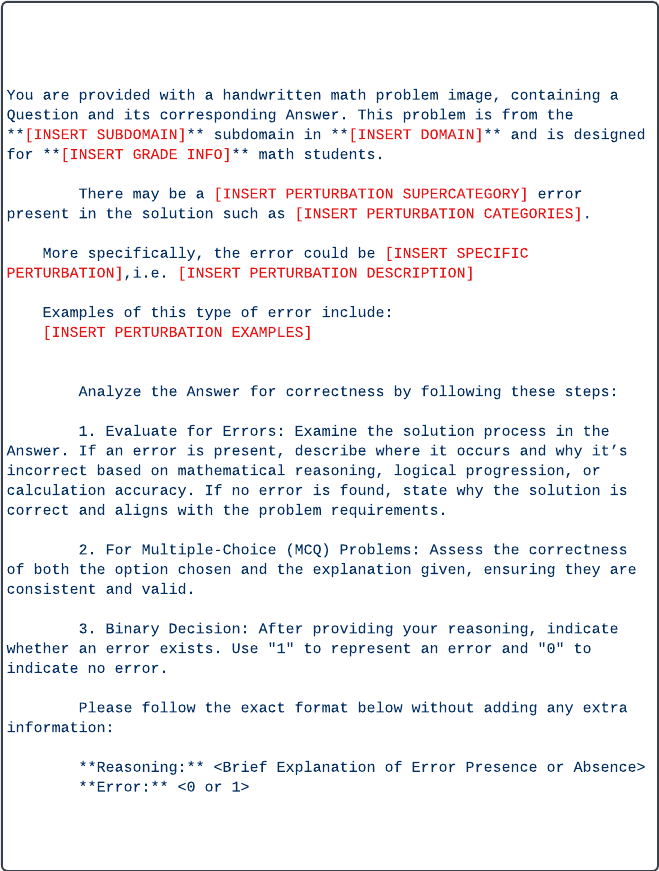}
    \caption{Level 3 Error Detection Prompt for \Cref{levels_expt}.}
    \label{fig:L3_prompt}
\end{figure*}

\begin{figure*}[h!]
    \centering
    \includegraphics[width=0.7\linewidth]{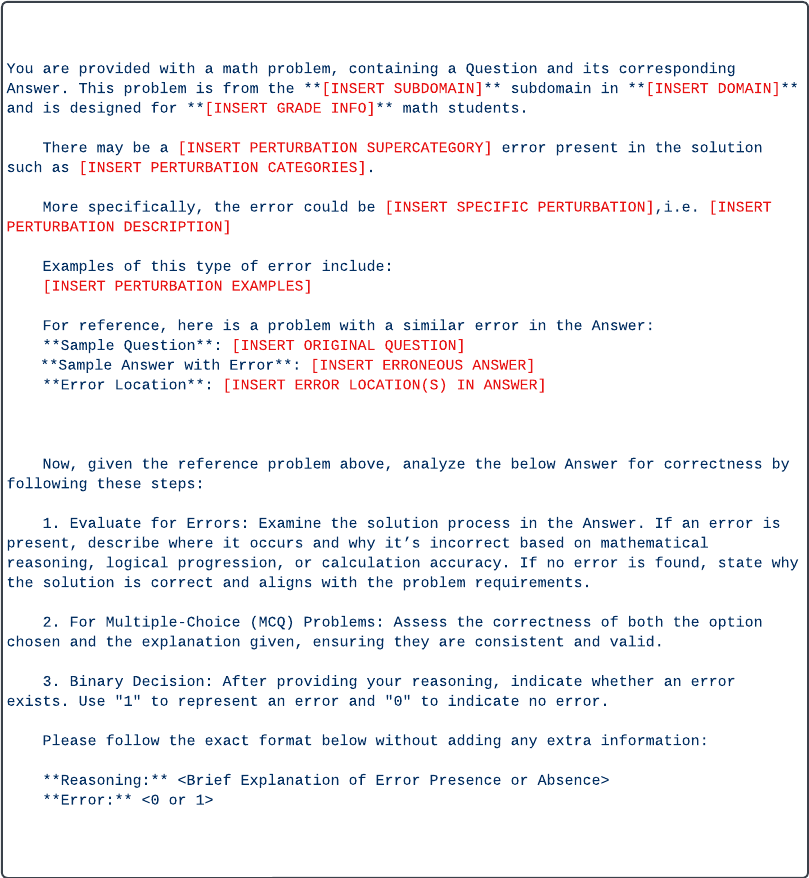}
    \caption{Level 4 Error Detection Prompt for \Cref{levels_expt}.}
    \label{fig:L4_prompt}
\end{figure*}



\end{document}